%% file: main.tex
\documentclass{article}

    \usepackage[final]{neurips_2021}

\usepackage[utf8]{inputenc} %
\usepackage[T1]{fontenc}    %
\usepackage{hyperref}       %
\usepackage{url}            %
\usepackage{booktabs}       %
\usepackage{amsfonts}       %
\usepackage{nicefrac}       %
\usepackage{microtype}      %
\usepackage{xcolor}         %

\hypersetup{
    colorlinks,
    linkcolor={blue!50!black},
    citecolor={blue!50!black},
    urlcolor={blue!80!black}
}

\input{subtex/package}

\title{Second-Order Neural ODE Optimizer}

\author{%
  Guan-Horng Liu, Tianrong Chen, Evangelos A. Theodorou\\
  Georgia Institute of Technology, USA\\
  \texttt{\{ghliu, tianrong.chen, evangelos.theodorou\}@gatech.edu}\\
}

\begin{document}

\maketitle

\begin{abstract}

  We propose a novel second-order optimization framework
  for training the emerging deep continuous-time models, specifically the
  Neural Ordinary Differential Equations (Neural ODEs).
  Since their training already involves expensive gradient computation by solving a backward ODE,
  deriving efficient second-order methods becomes highly nontrivial.
  Nevertheless,
  inspired by the recent Optimal Control (OC) interpretation of training deep networks,
  we show that a specific continuous-time OC methodology, called \textit{Differential Programming},
  can be adopted to derive backward ODEs for {higher-order} derivatives at the same $\calO(1)$ memory cost.
  We further explore a low-rank representation of the second-order derivatives and show that it leads to efficient preconditioned updates with the aid of Kronecker-based factorization.
  The resulting method -- named \textbf{SNOpt} -- converges much faster than first-order baselines in wall-clock time,
  and the improvement remains consistent
  across various applications,
  \eg image classification, generative flow, and time-series prediction.
  Our framework also enables direct architecture optimization, such as the integration time of Neural ODEs,
  with second-order feedback policies,
  strengthening the OC perspective
   as a principled tool of analyzing optimization in deep learning.
  Our code is available at \url{https://github.com/ghliu/snopt}.

\end{abstract}

\section{Introduction} \label{sec:1}
\input{subtex/01intro}
\pagebreak

\section{Preliminaries} \label{sec:2}
\input{subtex/02prelim}

\section{Approach} \label{sec:3}

\input{subtex/03approach1}

\input{subtex/04approach2}

\vspace{-2pt}

\section{Experiments} \label{sec:4}

\input{subtex/05experiment}

\vspace{-2pt}

\section{Conclusion} \label{sec:6}

\vspace{-2pt}

We present an efficient higher-order optimization framework for training Neural ODEs.
Our method -- named \textbf{SNOpt} -- differs from existing second-order methods in various aspects.
While it leverages similar factorization inherited in Kronecker-based methods \citep{martens2015optimizing},
the two methodologies differ fundamentally in that
we construct analytic ODE expressions for higher-order derivatives (Theorem~\ref{prop:1})
and compute them through \texttt{\markblue{ODESolve}}.
This retains the favorable $\calO(1)$ memory as opposed to their $\calO({T})$.
It also enables a flexible rank-based factorization in Proposition~\ref{prop:2}.
Meanwhile,
our method extends the recent trend of OCP-inspired methods \citep{li2017maximum,liu2021dynamic}
to deep continuous-time models, yet using a rather straightforward framework
without imposing additional assumptions, such as Markovian or game transformation.
To summarize,
our work advances several methodologies to the emerging deep continuous-time models, achieving strong empirical results and
opening up new opportunities for analyzing models such as Neural SDEs/PDEs.

\newpage

\begin{ack}
  The authors would like to thank Chia-Wen Kuo and Chen-Hsuan Lin for the meticulous proofreading,
  and Keuntaek Lee for providing additional computational resources.
  Guan-Horng Liu was supported by CPS NSF Award \#1932068, and Tianrong Chen was supported by ARO Award \#W911NF2010151.
\end{ack}

\bibliographystyle{icml2021}
\bibliography{reference.bib}

\newpage

\appendix

\section{Appendix}

\subsection{Review of Optimal Control Programming (OCP) Perspective of Training Discrete DNNs and Continuous-time OCP} \label{app:1}
\input{subtex/07appendix}

\subsection{Missing Derivations and Discussions in Section \ref{sec:3.1} and \ref{sec:3.2}} \label{app:2}
\input{subtex/08appendix2}

\subsection{Discussion on the Free-Horizon Optimization in Section~\ref{sec:3.3}} \label{app:3}
\input{subtex/09appendix3}

\subsection{Experiment Details} \label{app:4}

\input{subtex/10appendix4}

\subsection{Additional Experiments} \label{app:5}

\input{subtex/11appendix5}

\end{document}

%% file: subtex/package.tex
\input{subtex/math_commands.tex}

\input{subtex/math_util.tex}

\newcommand{\eg}{{\ignorespaces\emph{e.g.}}{ }}
\newcommand{\ie}{{\ignorespaces\emph{i.e.}}{ }}

\usepackage{mathtools}
\usepackage{tabularx}
\usepackage{framed}

\usepackage{amssymb}
\usepackage{amsthm}
\usepackage{multirow}

\newtheorem{theorem}{Theorem}

\newtheorem{proposition}[theorem]{Proposition}
\newtheorem{corollary}[theorem]{Corollary}

\newtheorem{remark}[theorem]{Remark}

\newcommand\numberthis{\addtocounter{equation}{1}\tag{\theequation}}
\usepackage{cancel}
\usepackage{lipsum}

\usepackage{capt-of}
\usepackage{wrapfig}

\usepackage{xcolor}
\usepackage{color,soul}
\colorlet{color1}{green!50!black}
\colorlet{color2}{orange!95!black}
\colorlet{color3}{red!80!black}
\colorlet{color4}{red!65!black}
\colorlet{color5}{blue!75!green}
\colorlet{blueee}{blue!50!black}

\definecolor{amaranth}{rgb}{0.9, 0.17, 0.31}

\newcommand{\markgreen}[1]{{\ignorespaces\color{color1} #1}}
\newcommand{\markblue}[1]{{\ignorespaces\color{color5} #1}}

\usepackage{footnote}

\let\oldsqrt\sqrt
\def\sqrt{\mathpalette\DHLhksqrt}
\def\DHLhksqrt#1#2{\setbox0=\hbox{$#1\oldsqrt{#2\,}$}\dimen0=\ht0
\advance\dimen0-0.2\ht0
\setbox2=\hbox{\vrule height\ht0 depth -\dimen0}%
{\box0\lower0.4pt\box2}}

\newcommand{\specialcell}[2][c]{%
  \begin{tabular}[#1]{@{}c@{}}#2\end{tabular}}

\newcommand{\specialcelll}[2][l]{%
  \begin{tabular}[#1]{@{}l@{}}#2\end{tabular}}

\usepackage{enumitem}
\usepackage{subfig}
\usepackage{arydshln}

\usepackage{algorithm}
\usepackage{algorithmic}
\newcommand*\mystrut[1]{\vrule width0pt height0pt depth#1\relax}

\usepackage{booktabs}       %

\usepackage{enumitem}
\usepackage{tikz}

\usepackage{empheq}
\newcommand*\widefbox[1]{\fbox{\hspace{1em}#1\hspace{1em}}}
\usepackage{tablefootnote}

\usepackage{textcomp}
\usepackage{colortbl}

\newcommand{\corcmidrule}[1][1pt]{%
  \\[\dimexpr-\normalbaselineskip-\belowrulesep-\aboverulesep-#1\relax]%
}

%% file: subtex/math_commands.tex
\usepackage{amsmath,amsfonts,bm}

\def\eqref#1{equation~\ref{#1}}

\def\1{\bm{1}}

\def\rd{{\textnormal{d}}}

\def\rva{{\mathbf{a}}}

\def\rvg{{\mathbf{g}}}
\def\rvh{{\mathbf{h}}}
\def\rvu{{\mathbf{i}}}

\def\rvp{{\mathbf{p}}}
\def\rvq{{\mathbf{q}}}

\def\rvu{{\mathbf{u}}}

\def\rvx{{\mathbf{x}}}

\def\rvz{{\mathbf{z}}}

\def\vb{{\bm{b}}}

\def\vu{{\bm{u}}}

\def\vx{{\bm{x}}}
\def\vy{{\bm{y}}}
\def\vz{{\bm{z}}}

\def\mA{{\bm{A}}}
\def\mB{{\bm{B}}}
\def\mC{{\bm{C}}}
\def\mD{{\bm{D}}}

\def\mI{{\bm{I}}}

\def\mS{{\bm{S}}}

\def\mU{{\bm{U}}}

\def\mW{{\bm{W}}}
\def\mX{{\bm{X}}}

\DeclareMathAlphabet{\mathsfit}{\encodingdefault}{\sfdefault}{m}{sl}
\SetMathAlphabet{\mathsfit}{bold}{\encodingdefault}{\sfdefault}{bx}{n}

\def\tQ{{\tens{Q}}}

\newcommand{\vectorize}{\mathrm{vec}}

%% file: subtex/math_util.tex
\def\comma{{ \text{ ,} }}
\def\period{{ \text{ .} }}

\def\intT{{ \int_{t_0}^{t_1} }}
\def\rintT{{ \int^{t_0}_{t_1} }}

\def\hvx{{ \bar{\vx} }}
\def\hvu{{ \bar{\vu} }}

\def\drvx{{\delta \rvx}}
\def\drvu{{\delta \rvu}}
\def\drvxt{{\delta \rvx_t}}
\def\drvut{{\delta \rvu_t}}
\def\Lx{{\ell_{\hvx} }}
\def\Lu{{\ell_{\hvu} }}
\def\Lxx{{{\ell}_{\hvx \hvx}}}
\def\Luu{{{\ell}_{\hvu \hvu}}}
\def\Lux{{{\ell}_{\hvu \hvx}}}
\def\Lxu{{{\ell}_{\hvx \hvu}}}

\def\hF{{\bar{F}}}

\def\Fu{{{F}_\hvu}}
\def\Fx{{{F}_\hvx}}
\def\FxT{{{F}_\hvx^\T}}
\def\FuT{{{F}_\hvu^\T}}

\def\Qx{{Q_{\hvx}}}
\def\Qu{{Q_{\hvu}}}
\def\Qxu{{Q_{\hvx \hvu}}}
\def\Qxx{{Q_{\hvx \hvx}}}
\def\Quu{{Q_{\hvu \hvu}}}
\def\Qux{{Q_{\hvu \hvx}}}

\newcommand{\norm}[1]{\left\lVert#1\right\rVert}

\def\Inv{{-1}}

\def\cspace{{\mathbb{R}^{m}}}

\def\transpose{{\mathsf{T}}}

\def\dt{{ \mathrm{d} t }}

\def\calD{{\cal D}}

\def\calL{{\cal L}}

\def\calO{{\cal O}}

\newcommand{\nicefracpartial}[2]{\nicefrac{\partial #1}{\partial  #2}}

\newcommand{\fracpartial}[2]{\frac{\partial #1}{\partial  #2}}
\newcommand{\fracdiff}[2]{\frac{\rd #1}{\rd  #2}}
\newcommand{\br}[1]{\left[#1\right]}
\newcommand{\pr}[1]{\left(#1\right)}
\newcommand{\T}{\mathsf{T}}
\newcommand*\bvec[1]{\begin{bmatrix}#1\end{bmatrix}}
\newcommand{\bmat}[4]{\begin{bmatrix} #1 & #2 \\ #3 & #4\end{bmatrix}}
\def\vec{{\mathrm{vec}}}

%% file: subtex/01intro.tex
\begin{figure}[h]
\vskip -0.15in
\begin{center}
\includegraphics[height=2.5cm]{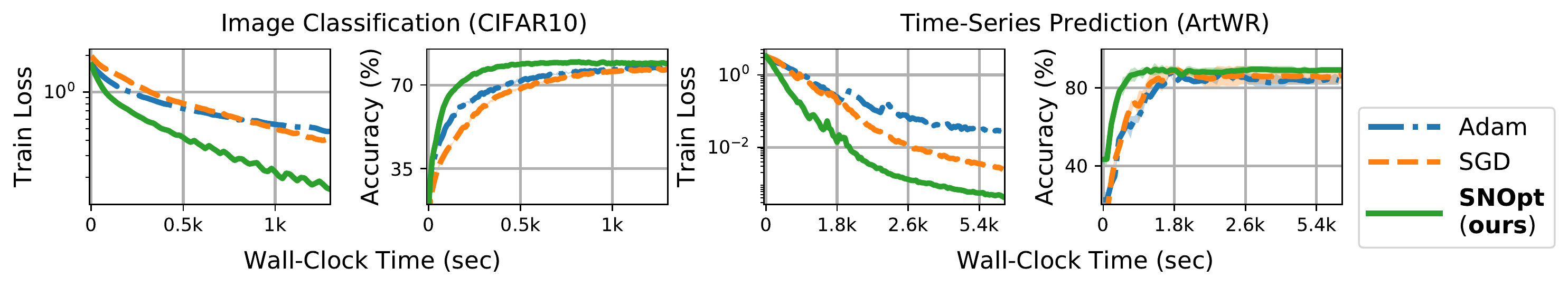}
\vskip -0.05in
\caption{
    Our second-order method ({SNOpt}; green solid curves) achieves superior convergence compared to first-order methods (SGD, Adam)
    on various Neural-ODE applications.
}
\label{fig:1}
\end{center}
\vskip -0.1in
\end{figure}

Neural ODEs \citep{chen2018neural} have received tremendous attention over recent years.
Inspired by taking the continuous limit of the ``discrete'' residual transformation,
$\rvx_{k+1} =  \rvx_k + \epsilon F(\rvx_k, \theta)$,
they propose to directly parameterize the vector field of an ODE as a deep neural network (DNN), \ie
\begin{align}
    \fracdiff{\rvx(t)}{t} = F(t, \rvx(t), \theta) , \quad  \rvx({t_0}) = \vx_{t_0},
    \label{eq:node}
\end{align}
where $\rvx(t) \in \mathbb{R}^m$ and $F(\cdot, \cdot, \theta)$ is a DNN parameterized by $\theta \in \mathbb{R}^n$.
This provides a {powerful} paradigm
connecting modern machine learning to classical differential equations \citep{weinan2017proposal}
and has since then achieved promising results {on} time series analysis \citep{rubanova2019latent,kidger2020neural}, reversible generative flow \citep{grathwohl2018ffjord,nguyen2019infocnf}, image classification \citep{zhuang2020adaptive,zhuang2021mali},
and manifold learning \citep{lou2020neural,mathieu2020riemannian}.

\begin{figure}[t]
\vskip -0.15in
\begin{center}
\setlength{\unitlength}{\textwidth}
\begin{picture}(1,0.15)
    \def\arraystretch{1.4}
    \put(0.03,0){\includegraphics[height=2.2cm]{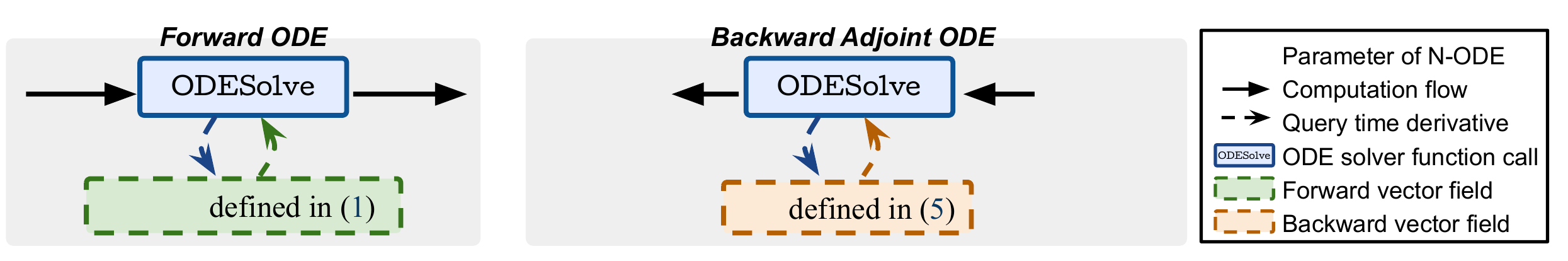}}
    \put(0.35,0.067){{\fontsize{9}{10}\selectfont$\left[\begin{array}{c}  $\quad\text{ }\text{ }$ \\ $\quad\text{ }\text{ }$ \\ $\quad\text{ }\text{ }$ \end{array}\right]$}}
    \put(0.355,0.067){{\fontsize{8.5}{10}\selectfont$\begin{array}{c} \vx(t_0) \\ \fracpartial{\calL}{\rvx(t_0)} \\ \fracpartial{\calL}{\theta} \end{array}$}}
    \put(0.66,0.067){{\fontsize{9}{10}\selectfont$\left[\begin{array}{c}  $\quad\text{ }\text{ }$ \\ $\quad\text{ }\text{ }$ \\ $\quad\text{ }\text{ }$ \end{array}\right]$}}
    \put(0.665,0.067){{\fontsize{8.5}{10}\selectfont$\begin{array}{c} \vx(t_1) \\ \fracpartial{\calL}{\rvx(t_1)} \\ \mathbf{0} \end{array}$}}
    \put(0.48 ,  0.026){{\fontsize{8.5}{10}\selectfont$G$}}
    \put(0.085 ,  0.027){{\fontsize{7}{10}\selectfont$F({\cdot},{\cdot},{\theta})$}}
    \put(0.075, 0.065){{\fontsize{7}{9}\selectfont$(t{,}\rvx(t))$}}
    \put(0.21, 0.065){{\fontsize{7}{9}\selectfont$\rd\rvx(t)/\dt$}}
    \put(0.045 , 0.11){{\fontsize{7}{10}\selectfont$\rvx(t_0)$}}
    \put(0.25 , 0.11){{\fontsize{7}{10}\selectfont$\rvx(t_1)$}}
    \put(0.78 , 0.118){{\fontsize{6}{10}\selectfont$\theta$}}
    \put(0.776 , 0.04){{\fontsize{4.5}{10}\selectfont$F$}}
    \put(0.776 , 0.02){{\fontsize{4.5}{10}\selectfont$G$}}
\end{picture}
\vskip -0.05in
\caption{
    Neural ODE features distinct training process:
    Both forward and backward passes parameterize
    vector fields
    so that any generic ODE solver (which can be non-differentiable) can query time derivatives,
    \eg $\fracdiff{\rvx(t)}{t}$,
    to solve the ODEs (\ref{eq:node}, \ref{eq:backward-ode}).
    In this work, we extend it to second-order training.
}
\label{fig:2}
\end{center}
\vskip -0.2in
\end{figure}

Due to the continuous-time representation,
Neural ODEs feature a distinct optimization process (see Fig.~\ref{fig:2}) compared to their discrete-time counterparts,
which also poses new challenges.
First, the forward pass of Neural ODEs involves solving (\ref{eq:node}) with a black-box ODE solver.
Depending on how its numerical integration is set up,
{the propagation may be refined to arbitrarily small step sizes and become prohibitively expensive to solve}
without any regularization \citep{ghosh2020steer,finlay2020train}.
On the other hand,
to prevent Back-propagating through the entire ODE solver,
the gradients are typically obtained by solving a \emph{backward adjoint ODE} using the Adjoint Sensitivity Method (ASM; \cite{pontryagin1962mathematical}).
While this can be achieved at a favorable $\calO(1)$ memory,
it further increases the runtime
and can suffer from inaccurate integration \citep{gholami2019anode}.
For these reasons,
Neural ODEs often take notoriously longer time to train,
limiting
their applications to relatively small or synthetic datasets \citep{massaroli2020dissecting} until very recently \citep{zhuang2021mali}.

To improve the convergence rate of training,
it is natural to consider higher-order optimization.
While efficient second-order methods have been proposed for discrete models \citep{ba2016distributed,george2018fast},
it remains unclear how to extend these successes to Neural ODEs, given their distinct computation processes.
Indeed, limited discussions in this regard only note that one may
repeat the backward adjoint process recursively to obtain higher-order derivatives \citep{chen2018neural}.
This is, unfortunately, impractical as the recursion will accumulate the aforementioned integration errors and scale the per-iteration runtime linearly.
As such,
second-order methods for Neural ODEs are seldom considered in practice, nor have they been rigorously explored from an optimization standpoint.

In this work, we show that efficient second-order optimization is in fact viable for Neural ODEs.
Our method is inspired by the emerging {Optimal Control perspective} \citep{han2018mean,liu2019deep}, which treats the parameter $\theta$ as a control variable,
so that the training process, \ie optimizing $\theta$ w.r.t. some objective, can be interpreted as an Optimal Control Programming (OCP).
Specifically,
we show that a continuous-time OCP methodology, called \emph{Differential Programming},
provides analytic second-order derivatives
by solving a set of {coupled matrix ODEs}.
Interestingly, these matrix ODEs can be augmented to the backward adjoint ODE and solved simultaneously.
In other words, a single backward pass is sufficient to compute all derivatives, including the original ASM-based gradient, the newly-derived second-order matrices, or even higher-order tensors.
{
Further, these higher-order computations enjoy the same $\calO(1)$ memory
and a comparable runtime to first-order methods
by adopting Kronecker factorization \citep{martens2015optimizing}.
The resulting method -- called \textbf{SNOpt} -- admits superior convergence in wall-clock time (Fig.~\ref{fig:1}),
and the improvement remains consistent across image classification, continuous normalizing flow, and time-series prediction.

Our OCP framework also facilitates progressive training of the network architecture.
Specifically, we study an example of jointly optimizing the ``integration time'' of Neural ODEs, in analogy to the ``depth'' of discrete DNNs.
While analytic gradients w.r.t.
this architectural parameter
have been derived under the ASM framework,
they were often evaluated on limited synthetic datasets \citep{massaroli2020dissecting}.
In the context of OCP, however,
free-horizon optimization
is a well-studied problem
for practical applications with a priori unknown terminal time \citep{sun2015model,de2019free}.
In this work, we
show that these principles can be applied to Neural ODEs,
yielding
a novel second-order \emph{feedback policy} that adapts the integration time throughout training.
On training CIFAR10, this further leads to a 20\% runtime reduction, yet without hindering test-time accuracy.
}

In summary, we present the following contributions.
\begin{itemize}[leftmargin=15pt]
\item
We propose a novel computational framework for computing higher-order derivatives of deep continuous-time models, with a rigorous analysis using continuous-time Optimal Control theory.
\item
We propose an efficient second-order method, \textbf{SNOpt}, that achieves
superior convergence (in wall-clock time) over first-order methods in training Neural ODEs,
while retaining constant memory complexity.
These improvements remain consistent across various applications.
\item
To show that our framework also enables direct architecture optimization,
we derive a second-order feedback policy for adapting the integration horizon and show it {further reduces the runtime}.
\end{itemize}

%% file: subtex/02prelim.tex
\textbf{Notation.}
We use roman and italic type to represent a variable $\rvx(t)$ and its realization $\vx(t)$ given an ODE.
\markblue{\texttt{ODESolve}} denotes a function call that solves an initial value problem given
an initial condition, start and end integration time, and vector field,
\ie \texttt{\markblue{ODESolve}(}$\rvx({t_0}), t_0, t_1, F$\texttt{)}
where $\fracdiff{\rvx(t)}{t} = F$.

\paragraph{Forward and backward computations of Neural ODEs.}
Given an initial condition $\rvx({t_0})$ and integration interval $[t_0, t_1]$,
Neural ODEs concern the following optimization over an objective $\calL$,
\begin{align}
\min_\theta \calL(\rvx(t_1)), \quad \text{where }
\rvx(t_1) = \rvx({t_0}) + \intT F(t, \rvx(t), \theta) \text{ } \dt
\label{eq:node-prob}
\end{align}
is the solution of the ODE (\ref{eq:node})
and can be solved by calling a black-box ODE solver, \ie
$\rvx(t_1) =$ \texttt{\markblue{ODESolve}(}$\rvx({t_0}), t_0, t_1, F$\texttt{)}.
The use of \markblue{\texttt{ODESolve}} allows us to adopt higher-order numerical methods,
\eg adaptive Runge-Kutta \citep{press2007numerical},
which give more accurate integration compared with \eg vanilla Euler discretization in residual-based discrete models.
To obtain the gradient $\fracpartial{\calL}{\theta}$ of Neural ODE,
one may naively Back-propagate through \markblue{\texttt{ODESolve}}.
This, even if it could be made possible,
leads to unsatisfactory memory complexity
since the computation graph can grow arbitrarily large for adaptive ODE solvers.
Instead,
\citet{chen2018neural} proposed to apply the Adjoint Sensitivity Method (ASM),
which states that the gradient can be obtained through the following integration.
\begin{align}
\fracpartial{\calL}{\theta} = -\rintT \rva(t)^{\T} \fracpartial{F(t, \rvx(t), \theta)}{\theta} \text{ } \dt \comma
\label{eq:grad-node}
\end{align}
where $\rva(t) \in \mathbb{R}^m $ is referred to the \textit{adjoint} state whose dynamics obey a \textit{backward adjoint ODE},
\begin{align}
- \fracdiff{\rva(t)}{t} = \rva(t)^\T \fracpartial{F(t, \rvx(t), \theta)}{\rvx(t)},
 \quad
\rva(t_1) = \fracpartial{\calL}{\rvx(t_1)}
\period
\label{eq:adjoint-node}
\end{align}
Equations (\ref{eq:grad-node}, \ref{eq:adjoint-node}) present two coupled ODEs that can be viewed as
the continuous-time expression of the Back-propagation \citep{lecun1988theoretical}.
Algorithmically, they can be solved through another call of \texttt{\markblue{ODESolve}} (see Fig.~\ref{fig:2}) with an augmented dynamics $G$, \ie
\begin{align}
\bvec{ \rvx(t_0) \\[0.5ex] \rva(t_0) \\[0.5ex] \nicefracpartial{\calL}{\theta} } =
\texttt{\markblue{ODESolve}(}\bvec{ \rvx(t_1) \\[0.5ex] \rva(t_1) \\[0.5ex] \mathbf{0} }, t_1, t_0, G  \texttt{)}, \text{ where }
G\left(t, \bvec{ \rvx(t) \\[0.5ex] \rva(t) \\[0.5ex] \cdot }, \theta\right) := \bvec{
    F(t, \rvx(t), \theta) \\[0.5ex]
    -\rva(t)^\T \fracpartial{F}{\rvx} \\[0.5ex]
    -\rva(t)^\T \fracpartial{F}{\theta}}
\label{eq:backward-ode}
\end{align}
augments the original dynamics $F$ in (\ref{eq:node}) with the adjoint ODEs (\ref{eq:grad-node}, \ref{eq:adjoint-node}).
Notice that this computation~(\ref{eq:backward-ode}) depends only on $(\rvx(t_1),\rva(t_1))$.
This differs from naive Back-propagation,
which requires storing intermediate states along the entire computation graph of forward \markblue{\texttt{ODESolve}}.
While the latter requires {$\calO(\widetilde{T})$}
memory cost,\footnote{
    \smash{$\widetilde{T}$} is the number of the adaptive steps used to solve (\ref{eq:node}),
    as an analogy of the ``depth'' of Neural ODEs.
} the computation in (\ref{eq:backward-ode}) only consumes constant $\calO(1)$ memory cost.

\begin{minipage}{0.39\textwidth}
\citet{chen2018neural} noted that if we further encapsulate (\ref{eq:backward-ode}) by $\fracpartial{}{\theta}\calL = \texttt{{grad}(} \calL, \theta \texttt{)}$,
one may compute higher-order derivatives by recursively calling
$\fracpartial{^{n}\calL}{\theta^{n}} = \texttt{grad(} \fracpartial{^{n-1}\calL}{\theta^{n-1}}, \theta \texttt{)}$, starting from $n{=}1$.
This can scale unfavorably due to its recursive dependence and
accumulated integration errors.
Indeed, Table~\ref{table:adj2-error}
\end{minipage}
\hfill
\begin{minipage}{0.58\textwidth}
    \centering
    \captionsetup{type=table}
    \caption{
        Numerical errors
        between ground-truth and adjoint derivatives
        using
        different \markblue{\texttt{ODESolve}} on CIFAR10.
    }
    \vskip -0.03in
    \centering
    \begin{tabular}{c|ccc}
      \toprule
       & \texttt{rk4} & \texttt{implicit adams} & \texttt{dopri5} \\
      \midrule
      $\fracpartial{\calL}{\theta}$
      &   7.63$\times$10$^{-\text{5}}$ & 2.11$\times$10$^{-\text{3}}$ & 3.44$\times$10$^{-\text{4}}$ \\[3pt]
      $\fracpartial{^2\calL}{\theta^2}$
      & 6.84$\times$10$^{-\text{3}}$   & 2.50$\times$10$^{-\text{1}}$    & 41.10 \\
      \bottomrule
    \end{tabular} \label{table:adj2-error}
\end{minipage}
suggests that
the errors of second-order derivatives, $\fracpartial{^2\calL}{\theta^2}$,
obtained from the recursive adjoint procedure
can be 2-6 orders of magnitude
larger than the ones from the first-order adjoint, $\fracpartial{\calL}{\theta}$.
In the next section,
we will present a novel optimization framework that
computes these higher-order derivatives \textit{without} any recursion (Section~\ref{sec:3.1})
and discuss how it can be implemented efficiently (Section~\ref{sec:3.2}).

%% file: subtex/03approach1.tex
\subsection{Dynamics of Higher-order Derivatives using Continuous-time Optimal Control Theory} \label{sec:3.1}

\colorlet{ggreen}{green!60!black}
\colorlet{rred}{red!80!black}
\setul{0.2ex}{0.2ex}

OCP perspective is a recently emerging methodology for analyzing optimization of discrete DNNs.
Central to its interpretation is to treat the layer propagation of a DNN as discrete-time dynamics,
so that the training process, \ie finding an
\setulcolor{ggreen}
\ul{\textit{optimal parameter}} of
\setulcolor{rred}
\ul{\textit{a DNN}},
can be understood like an OCP, which searches for an
\setulcolor{ggreen}
\ul{\textit{optimal control}} subjected to
\setulcolor{rred}
\ul{\textit{a dynamical constraint}}.
This perspective
has provided useful insights on
characterizing the optimization process \citep{hu2019mean} and enhancing
principled algorithmic design \citep{liu2021differential}.
We leave a complete discussion in Appendix~{\ref{app:1}}.

Lifting this OCP perspective from discrete DNNs to Neural ODEs %
requires special treatments from continuous-time OCP theory \citep{todorov2006optimal}.
Nevertheless, we highlight that
training Neural ODEs and solving continuous-time OCP
are fundamentally intertwined since these models, by construction, represent continuous-time dynamical systems.
Indeed, the ASM used for deriving (\ref{eq:grad-node},~\ref{eq:adjoint-node}) originates from the celebrated Pontryagin's principle \citep{pontryagin1962mathematical},
which is an optimality condition to OCP.
Hence, OCP analysis is not only {motivated but principled} from an optimization standpoint.

We begin by first transforming (\ref{eq:node-prob}) to a form
that is easier to adopt the continuous-time OCP analysis.
\begin{equation}
\begin{split}
\min_\theta \br{\Phi(\rvx_{t_1}) + \intT \ell(t,\rvx_t, \rvu_t) \dt } \quad \text{subjected to }
\begin{cases}
\fracdiff{\rvx_t}{t} = F(t, \rvx_t, \rvu_t), \quad & \rvx_{t_0} = \vx_{t_0} \\
\fracdiff{\rvu_t}{t} = \mathbf{0}          , \quad & \rvu_{t_0} = \theta
\end{cases},
 \label{eq:ct-ocp}
\end{split}
\end{equation}
where $\rvx(t) \equiv \rvx_t$, and etc.
It should be clear that (\ref{eq:ct-ocp}) describes (\ref{eq:node-prob}) without loss of generality
by having $(\Phi, \ell) := (\calL, 0)$.
These functions are known as the terminal and intermediate costs in standard OCP.
In training Neural ODEs, $\ell$ can be used to describe either the weight decay, \ie $\ell \propto \norm{\rvu_t}$, or more complex regularization \citep{finlay2020train}.
The time-invariant ODE imposed for $\rvu_t$ makes the ODE of $\rvx_t$ equivalent to (\ref{eq:node}).
Problem (\ref{eq:ct-ocp}) shall be understood as
a particular type of OCP that searches for an optimal initial condition $\theta$ of a time-invariant control $\rvu_t$.
Despite seemly superfluous,
this is a necessary transformation that enables rigorous OCP analysis for the original training process (\ref{eq:node-prob}),
and it has also appeared in other control-related analyses \citep{zhong2020symplectic,chalvidal2021go}.

Next, define the accumulated loss from any time $t \in [t_0,t_1]$ to the integration end time $t_1$ as
\begin{align}
Q(t,\rvx_t,\rvu_t) := {\Phi(\rvx_{t_1})+ \int_t^{t_1} \ell(\tau,\rvx_\tau,\rvu_\tau) \text{ } \rd \tau },
\label{eq:Q}
\end{align}
which is also known in OCP as the \emph{cost-to-go} function.
Recall that our goal is to compute higher-order derivatives w.r.t. the parameter $\theta$ of Neural ODEs.
Under the new OCP representation (\ref{eq:ct-ocp}), the first-order derivative $\fracpartial{\calL}{\theta}$ is identical to
$\fracpartial{Q(t_0, \rvx_{t_0}, \rvu_{t_0})}{\rvu_{t_0}}$.
This is because $Q(t_0, \rvx_{t_0}, \rvu_{t_0})$ accumulates all sources of losses between $[t_0,t_1]$ (hence it sufficiently describes $\calL$)
and $\rvu_{t_0} = \theta$ by construction.
Likewise,
the second-order derivatives can be captured by the Hessian
$\fracpartial{^2 Q(t_0, \rvx_{t_0}, \rvu_{t_0})}{\rvu_{t_0}\partial\rvu_{t_0}} = \fracpartial{^2\calL}{\theta\partial\theta} \equiv \calL_{\theta\theta}$.
In other words,
we are only interested in obtaining the derivatives of $Q$ at the integration start time $t_0$.

To obtain these derivatives,
notice that we can rewrite (\ref{eq:Q}) as
\begin{align}
{0 = {\ell(t,\rvx_t,\rvu_t) + \fracdiff{Q(t,\rvx_t,\rvu_t)}{t}}}, \quad
Q({t_1},\rvx_{t_1}) = \Phi(\rvx_{t_1}),
\label{eq:Q-ode}
\end{align}
since the definition of $Q$ implies that
$Q(t,\rvx_t,\rvu_t) ={\ell(t,\rvx_t,\rvu_t) \dt + Q(t+\dt,\rvx_{t+\dt},\rvu_{t+\dt})}$.
We now state our main result,
which provides a local characterization of (\ref{eq:Q-ode}) with a set of coupled ODEs expanded along a solution path.
These ODEs can be used to obtain {all second-order derivatives at $t_0$}.
\begin{theorem}[Second-order Differential Programming] \label{prop:1}
Consider a solution path $(\hvx_t, \hvu_t)$
that solves the ODEs in (\ref{eq:ct-ocp}).
Then the first and second-order derivatives of $Q(t,\rvx_t,\rvu_t)$,
expanded locally around this solution path,
obey the following backward ODEs:
\begin{subequations}
\begin{alignat}{2}
- \fracdiff{\Qx}{t} &= \Lx + \FxT\Qx, \quad
&&\text{ }\text{ }- \fracdiff{\Qu}{t} = \Lu + \FuT\Qx, \label{eq:cddp-qa} \\
- \fracdiff{\Qxx}{t} &= \Lxx + \FxT\Qxx+\Qxx\Fx, \quad
&&- \fracdiff{\Qxu}{t} = \Lxu + \Qxx\Fu + \FxT\Qxu, \label{eq:cddp-qb} \\
- \fracdiff{\Quu}{t} &= \Luu + \FuT\Qxu + \Qux\Fu, \quad
&&- \fracdiff{\Qux}{t} = \Lux + \FuT\Qxx + \Qux\Fx, \label{eq:cddp-qc}
\end{alignat} \label{eq:cddp-q}
\end{subequations}
where
$\Fx(t) {\equiv} \fracpartial{F}{\rvx_t}|_{(\hvx_t,\hvu_t)}$,
$\Qxx(t) {\equiv} \fracpartial{^2Q}{\rvx_t\partial\rvx_t}|_{(\hvx_t,\hvu_t)}$, and etc.
All terms in (\ref{eq:cddp-q})
are time-varying vector-valued or matrix-valued functions expanded at $(\hvx_t, \hvu_t)$.
The terminal condition is given by
\begin{align*}
\Qx(t_1) = \Phi_\hvx, \quad
\Qxx(t_1) = \Phi_{\hvx\hvx}, \quad \text{ and } \quad
\Qu(t_1) = \Quu(t_1) = \Qux(t_1) = \Qxu(t_1) = \mathbf{0}.
\end{align*}
\end{theorem}

The proof (see Appendix~\ref{app:2}) relies on rewriting (\ref{eq:Q-ode})
with \emph{differential states},
$\drvx_t := \rvx_t - \hvx_t$,
which view the deviation from $\hvx_t$ as an \emph{optimizing variable}
(hence the name ``\emph{Differential Programming}'').
It can be shown that $\drvx_t$ follows a linear ODE expanded along the solution path.
Theorem \ref{prop:1} has several important implications.
First, the ODEs in (\ref{eq:cddp-qa}) recover the original ASM computation (\ref{eq:grad-node},\ref{eq:adjoint-node}),
as one can readily verify that
$\Qx(t) \equiv \rva(t)$ follows the same backward ODE in~(\ref{eq:adjoint-node})
and the solution of the second ODE in (\ref{eq:cddp-qa}), $Q_{\hvu}(t_0) = - \rintT \Fu^\T\Qx \dt $,
gives the exact gradient in~(\ref{eq:grad-node}).
Meanwhile,
solving the coupled matrix ODEs presented in (\ref{eq:cddp-qb}, \ref{eq:cddp-qc})
yields the desired second-order matrix, $\Quu(t_0) \equiv \calL_{\theta\theta}$, for preconditioning the update.
Finally,
one can derive the dynamics of other higher-order tensors using the same Differential Programming methodology
by simply expanding (\ref{eq:Q-ode}) beyond the second order.
We leave some discussions in this regard in Appendix~\ref{app:2}.

%% file: subtex/04approach2.tex
\definecolor{darkgreen}{HTML}{38761D}

\subsection{Efficient Second-order Preconditioned Update} \label{sec:3.2}

Theorem~\ref{prop:1} provides an attractive computational framework
that does not require recursive computation (as mentioned in Section~\ref{sec:2}) to obtain higher-order derivatives.
It suggests that
we can obtain first and second-order derivatives all at once with a single function call of \texttt{\markblue{ODESolve}}:
\begin{align}
\begin{split}
[\vx_{t_0}, \Qx(t_0), \Qu(t_0), \Qxx(t_0), &\text{ }\Qux(t_0), \Qxu(t_0), \Quu(t_0)]  \\ &=
\texttt{\markblue{ODESolve}(}[\vx_{t_1}, \Phi_\hvx, \mathbf{0}, \Phi_{\hvx\hvx}, \mathbf{0}, \mathbf{0}, \mathbf{0}], t_1, t_0, \tilde{G} \texttt{)},
\label{eq:backward-ode2}
\end{split}
\end{align}
where $\tilde{G}$ augments the original dynamics $F$ in (\ref{eq:node}) with all 6 ODEs presented in (\ref{eq:cddp-q}).
Despite that this OCP-theoretic backward pass (\ref{eq:backward-ode2}) retains the same $\calO(1)$ memory complexity as in (\ref{eq:backward-ode}),
the dimension of the new augmented state, which now carries second-order matrices,
can grow to an unfavorable size that dramatically slows down the numerical integration.
Hence, we must consider other representations of (\ref{eq:cddp-q}), if any, in order to proceed.
In the following proposition, we present one of which
that transforms (\ref{eq:cddp-q}) into a set of {vector} ODEs, so that we can compute them much efficiently.
\begin{proposition}[Low-rank representation of (\ref{eq:cddp-q})] \label{prop:2}
    Suppose $\ell {:=} 0$ in (\ref{eq:ct-ocp})
    and let $\Qxx(t_1){ =} \sum_{i=1}^R \vy_i \otimes \vy_i$ be a symmetric matrix of rank $R \le n$,
    where $\vy_i \in \mathbb{R}^m$ and $\otimes$ is the Kronecker product.
    Then, for all $t \in [t_0, t_1]$, the second-order matrices appeared in (\ref{eq:cddp-qb}, \ref{eq:cddp-qc}) can be decomposed into
    \begin{align*}
        \Qxx(t) = \sum_{i=1}^R \rvq_i(t) \otimes \rvq_i(t), \quad
        \Qxu(t) = \sum_{i=1}^R \rvq_i(t) \otimes \rvp_i(t), \quad
        \Quu(t) = \sum_{i=1}^R \rvp_i(t) \otimes \rvp_i(t),
    \end{align*}
    where the vectors $\rvq_i(t) \in \mathbb{R}^m $ and $\rvp_i(t) \in \mathbb{R}^n$ obey the following backward ODEs:
    \begin{align}
        - \fracdiff{\rvq_i(t)}{t} = \Fx(t)^\T\rvq_i(t), \quad
        - \fracdiff{\rvp_i(t)}{t} = \Fu(t)^\T\rvq_i(t), \quad
        \label{eq:vector-ode}
    \end{align}
    with the terminal condition given by $(\rvq_i({t_1}),\rvp_i({t_1})) := (\vy_i, \mathbf{0})$.
\end{proposition}
The proof is left in Appendix~\ref{app:2}.
Proposition~\ref{prop:2} gives a nontrivial conversion.
It indicates that the \emph{coupled matrix} ODEs presented in (\ref{eq:cddp-qb}, \ref{eq:cddp-qc}) can be disentangled into a set of \emph{independent vector} ODEs where each of them follows its
own dynamics (\ref{eq:vector-ode}).
As the rank $R$ determines the number of these vector ODEs,
this conversion will be particularly useful if the second-order matrices exhibit low-rank structures.
Fortunately, this is indeed the case for many Neural-ODE applications
which often propagate $\rvx_t$ in a latent space of higher dimension \citep{chen2018neural,grathwohl2018ffjord,kidger2020neural}.

Based on Proposition~\ref{prop:2}, the second-order precondition matrix $\calL_{\theta\theta}$
is given by\footnote{
    We drop the dependence on $t$ for brevity, yet all terms inside the integrations of (\ref{eq:Quu}, \ref{eq:q-kfac}) are time-varying.
}
\begin{align}
\calL_{\theta\theta} \equiv \Quu(t_0) =
\sum_{i=1}^R \left( \rintT  \Fu^\T\rvq_i \text{ } \dt \right) \otimes \left( \rintT  \Fu^\T\rvq_i \text{ } \dt \right),
\label{eq:Quu}
\end{align}
where $\rvq_i \equiv \rvq_i(t)$ follows (\ref{eq:vector-ode}).
Our final step is to facilitate efficient computation of (\ref{eq:Quu}) with Kronecker-based factorization,
which underlines many popular second-order methods for discrete DNNs \citep{grosse2016kronecker,martens2018kronecker}.
Recall that
the vector field $F$ is represented
\begin{wrapfigure}[7]{r}{0.4\textwidth}
  \vspace{-7pt}
  \setlength{\unitlength}{0.1\textwidth}
  \begin{center}
    \begin{picture}(4,1.5)
      \put(0,0){\includegraphics[height=2.12cm]{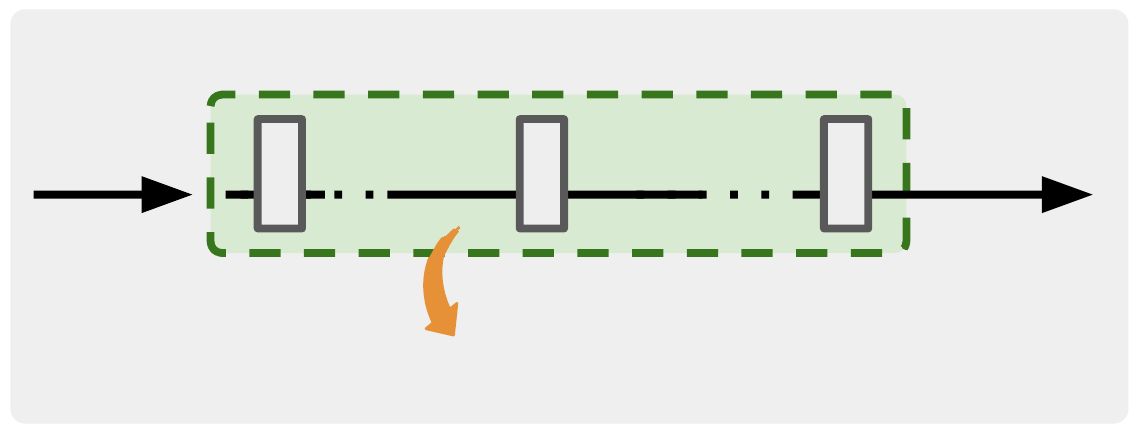}}
      \put(1.05, 1.27){{\color{darkgreen}\fontsize{8}{9}\selectfont$F({\cdot},{\cdot},{\theta}) \equiv F({\cdot},{\cdot},{\rvu_t}) $}}
      \put(1.3, 0.93){{\fontsize{7}{9}\selectfont$\rvz^n(t)$}}
      \put(2.05, 0.93){{\fontsize{7}{9}\selectfont$\rvz^{n{+}1}(t)$}}
      \put(0.05, 1){{\fontsize{7}{9}\selectfont$(t{,}\rvx(t))$}}
      \put(3.25, 1){{\fontsize{9.5}{10}\selectfont$\frac{\rd\rvx(t)}{\dt}$}}
      \put(1.58, 0.27){{\fontsize{6}{7}\selectfont$\begin{cases} $\text{ }$ \\ $\text{ }$ \end{cases}$}}
      \put(1.89, 0.37){{\fontsize{7.5}{9}\selectfont$\rvh^n(t) = f{(}\rvz^n{(}t{)}{,}\rvu^n{(}t{)}{)}$}}
      \put(1.7, 0.12){{\fontsize{7.5}{9}\selectfont$\rvz^{n{+}1}(t) = \sigma(\rvh^n(t))$}}
    \end{picture}
  \end{center}
  \vskip -0.1in
  \caption{The layer propagation inside the vector field $F$, where $f$ and $\sigma$ denote affine and nonlinear activation functions.}
  \label{fig:3}
\end{wrapfigure}
by a DNN.
Let $\rvz^n(t)$, $\rvh^n(t)$, and $\rvu^n(t)$ denote
the activation vector, pre-activation vector, and the parameter of layer $n$
when evaluating $\fracdiff{\rvx}{t}$ at time $t$ (see Fig.~\ref{fig:3}),
then the integration in (\ref{eq:Quu}) can be broken down into each layer $n$,
\begin{align*}
\rintT  \left(\Fu^\T\rvq_i\right) \dt
=& [\cdots, \textstyle \rintT  \left(F_{\hvu^n}^\T\rvq_i\right) \dt, \cdots] \\
=& [\cdots, \textstyle \rintT \left( \rvz^n \otimes ({\fracpartial{F}{\rvh^n}}^\T \rvq_i) \right) \dt, \cdots],
\end{align*}
where the second equality holds by
$
F_{\hvu^n}^\T \rvq_i
= (\fracpartial{F}{\rvh^n} \fracpartial{\rvh^n}{\rvu^n}) ^\T \rvq_i
= \rvz^n \otimes ({\fracpartial{F}{\rvh^n}}^\T \rvq_i)$.
This is an essential step towards the Kronecker approximation of the layer-wise precondition matrix:
\begin{align*}
\calL_{\theta^n\theta^n}
\equiv
Q_{\hvu^n \hvu^n}(t_0)
&=
\sum_{i=1}^R
\left( \rintT \left( \textstyle \rvz^n \otimes ({\fracpartial{F}{\rvh^n}}^\T \rvq_i) \right) \dt \right)
\otimes
\left( \rintT \left( \textstyle
\rvz^n \otimes ({\fracpartial{F}{\rvh^n}}^\T \rvq_i) \right) \dt \right) \\
&\approx
\rintT \underbrace{\mystrut{2.5ex} \left( \textstyle \rvz^n \otimes \rvz^n \right)}_{\mA_n(t)} \dt
\otimes
\rintT \underbrace{ \sum_{i=1}^R \left( \textstyle ({\fracpartial{F}{\rvh^n}}^\T \rvq_i) \otimes ({\fracpartial{F}{\rvh^n}}^\T \rvq_i) \right)}_{\mB_n(t)} \dt.
\numberthis \label{eq:q-kfac}
\end{align*}

\begin{figure}[t]
  \vskip -0.2in
  \begin{minipage}{\textwidth}
  \begin{algorithm}[H]
   \vskip -0.05in
   \caption{SNOpt: Second-order Neural ODE Optimizer} %
   \label{alg:1}
   \begin{algorithmic}[1]
   \STATE {\bfseries Input:}
        dataset $\calD$,
        parametrized vector field $F(\cdot,\cdot,\theta)$,
        integration time $[t_0,t_1]$,
        black-box ODE \\ $\quad$ solver \texttt{\markblue{ODESolve}},
        learning rate $\eta$,
        rank $R$, interval of the time grid $\Delta t$
   \REPEAT

   \STATE Solve $\rvx(t_1) =$ \texttt{\markblue{ODESolve}(}$\rvx({t_0}), t_0, t_1, F$\texttt{)},
          where $\rvx({t_0}) \sim \calD$. \hfill \markgreen{$\rhd$ Forward pass}
   \STATE Initialize $(\bar{\mA}_n,\bar{\mB}_n) := (\mathbf{0}, \mathbf{0})$ for each layer $n$
          and set $\rvq_i(t_1) := \vy_i$.
   \FOR{$t^\prime$ {\bfseries in} $\{t_1, t_1-\Delta t, \cdots, t_0+\Delta t, t_0 \}$ }
    \STATE
        Set $t := t^\prime - \Delta t$ as the small integration step, then call \\[0.3ex]
        $\quad [\rvx({{t}}), \Qx({t}), \Qu({t}), \{\rvq_i({t})\}_{i=1}^R]$ \\[0.3ex]
        $\quad\quad = \texttt{\markblue{ODESolve}(}[\rvx({t^\prime}), \Qx({t^\prime}), \Qu({t^\prime}), \{\rvq_i({t^\prime})\}_{i=1}^R ], t^\prime, t, \widehat{G} \texttt{)}$,
            \hfill \markgreen{$\rhd$ Backward pass\text{ }} \\
        where $\widehat{G}$ augments the ODEs of state (\ref{eq:node}), first and second-order derivatives (\ref{eq:cddp-qa}, \ref{eq:vector-ode}).
    \STATE Evaluate $\rvz^n(t)$, $\rvh^n(t)$, $F(t, \rvx_{{t}}, \theta)$, then
           compute $\mA_n(t), \mB_n(t)$ in (\ref{eq:q-kfac}).
    \STATE Update $\bar{\mA}_n \leftarrow \bar{\mA}_n + \mA_n(t) \cdot \Delta t $
           and    $\bar{\mB}_n \leftarrow \bar{\mB}_n + \mB_n(t) \cdot \Delta t $.
   \ENDFOR
   \STATE $\forall n$, apply
    $ \theta^n \leftarrow \theta^n - \eta \cdot \vec(\bar{\mB}_n^{-1}Q_{\hvu^n}(t_0)\bar{\mA}_n^{-\T})$.
    \hfill \markgreen{$\rhd$ Second-order parameter update\text{ }}
   \UNTIL{ converges }
   \end{algorithmic}
  \end{algorithm}
  \end{minipage}
  \vskip 0.05in
  \begin{minipage}{\textwidth}
      \centering
      \setlength{\unitlength}{\textwidth}
      \begin{picture}(1,0.175)
          \put(0.05,0){\includegraphics[height=2.5cm]{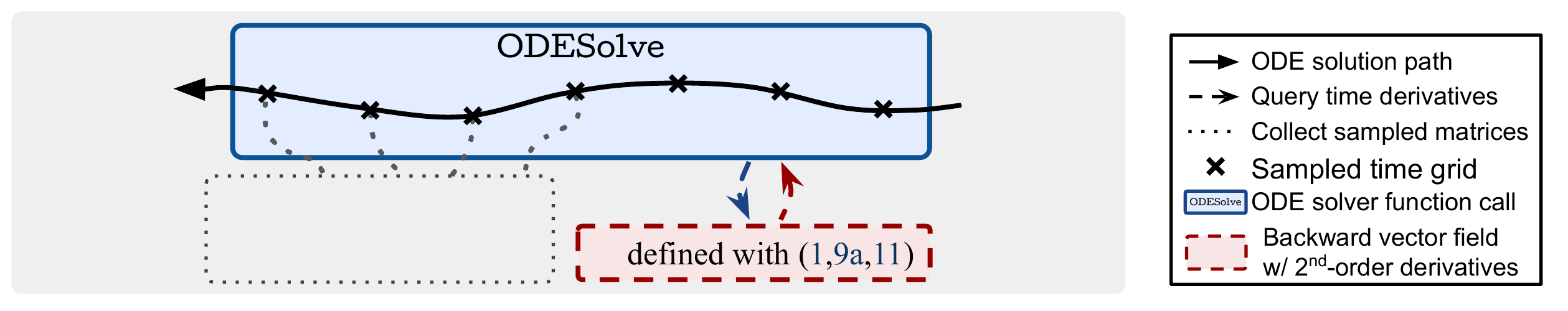}}
          \put(0.053,0.105){{\fontsize{9}{10}\selectfont$\left[\begin{array}{c}  $\quad\text{ }\text{ }$ \\ $\quad\text{ }\text{ }$ \\ $\quad\text{ }\text{ }$ \\ $\quad\text{ }\text{ }\text{ }$  \end{array}\right]$}}
          \put(0.054,0.105){{\fontsize{8.5}{10}\selectfont$\begin{array}{c}~\rvx(t_0) \\ Q_{\bar{\vx}}(t_0) \\  Q_{\bar{\vu}}(t_0) \\ \cdot~\end{array}$}}
          \put(0.598,0.105){{\fontsize{8.5}{10}\selectfont$\left[\begin{array}{c}  $\qquad$ \\ $\qquad$ \\ $\qquad$ \\ $\qquad$  \end{array}\right]$}}
          \put(0.598,0.105){{\fontsize{8.5}{10}\selectfont$\begin{array}{c}~\rvx(t_1) \\ Q_{\bar{\vx}}(t_1) \\  Q_{\bar{\vu}}(t_1) \\ \{\vy_i\}_{i=1}^R ~\end{array}$}}
          \put(0.31, 0.125){{\fontsize{8.5}{10}\selectfont$t_{j+1}$}}
          \put(0.39, 0.11){{\fontsize{8.5}{10}\selectfont$t_{j}$}}
          \put(0.901 , 0.078){{\fontsize{6}{10}\selectfont${\{}t_j{\}}$}}
          \put(0.17, 0.044){{\fontsize{7.5}{10}\selectfont$ \begin{cases} \bar{\mA}_n {=} {\sum_j} \mA_n(t_j) {\cdot} \Delta t \\ \bar{\mB}_n {=} {\sum_j} \mB_n(t_j) {\cdot} \Delta t \end{cases}$}}
          \put(0.387, 0.024){{\fontsize{8.5}{10}\selectfont$\widehat{G}$}}
          \put(0.7425, 0.027){{\fontsize{5}{10}\selectfont$\widehat{G}$}}
      \end{picture}
      \vskip -0.1in
      \caption{
        Our second-order method, SNOpt, solves a new backward ODE,
        \ie the $\widehat{G}$ appeared in line 6 of Alg.~\ref{alg:1},
        which augments second-order derivatives,
        while simultaneously collecting the matrices ${\mA}_n(t_j)$ and ${\mB}_n(t_j)$ on a sampled time grid $\{t_j\}$ for computing the
        preconditioned update in (\ref{eq:layer-kfac}).
      }
      \label{fig:4}
  \end{minipage}
  \vskip -0.05in
\end{figure}

\vspace{-6pt}

We discuss the approximation behind (\ref{eq:q-kfac}), and also the one for (\ref{eq:layer-kfac}), in Appendix~\ref{app:2}.
Note that $\mA_n(t)$ and $\mB_n(t)$ are much smaller matrices in $\mathbb{R}^{m\times m}$ compared to the ones in (\ref{eq:cddp-q}), and
they can be efficiently computed with
automatic differentiation packages \citep{paszke2017automatic}.
Now, let
$\{ t_j \}$ be a time grid uniformly distributed over $[t_0,t_1]$
so that
$\bar{\mA}_n {=} \textstyle \sum_j \mA_n(t_j) \Delta t$ and
$\bar{\mB}_n {=} \textstyle \sum_j \mB_n(t_j) \Delta t$ approximate
the integrations in (\ref{eq:q-kfac}),
then our final preconditioned update law is given by
\begin{align}
\forall n, \quad
\calL_{\theta^n\theta^n}^{-1} \calL_{\theta^n}
\approx \vec\left( \bar{\mB}_n^{-1} Q_{\hvu^n}(t_0) \bar{\mA}_n^{-\T} \right),
\label{eq:layer-kfac}
\end{align}
where $\vec$ denotes vectorization.
Our second-order method -- named \textbf{SNOpt} --  is summarized in Alg.~\ref{alg:1},
with the backward computation (\ie line 4-9 in Alg.~\ref{alg:1}) illustrated in Fig.~\ref{fig:4}.
In practice, we also adopt
eigen-based amortization with Tikhonov regularization (\citet{george2018fast}; see Alg.~\ref{alg:2} in Appendix~\ref{app:4}),
which stabilizes the updates over stochastic training.

\textbf{Remark.}
The fact that Proposition~\ref{prop:2} holds only for degenerate $\ell$
can be easily circumvented in practice.
As $\ell$ typically represents weight decay, $\ell := \frac{1}{t_1-t_0} \norm{\theta}_2$,
which is time-independent,
it can be separated from the backward ODEs (\ref{eq:cddp-q}) and added after solving the backward integration, \ie
\begin{align*}
\Qu(t_0) \leftarrow \gamma \theta + \Qu(t_0), \quad
\Quu(t_0)
\leftarrow \gamma \mI + \Quu(t_0),
\end{align*}
where $\gamma$ is the regularization factor.
Finally, we find that using the scaled \emph{Gaussian-Newton} matrix,
\ie $\Qxx(t_1) \approx \frac{1}{t_1 - t_0} \Phi_\hvx \otimes \Phi_\hvx$,
generally provides a good trade-off between the performance and runtime complexity.
As such, we adopt this approximation to Proposition~\ref{prop:2} for all experiments.

\subsection{Memory Complexity Analysis}

\begin{figure}[H]
  \centering
  \begin{minipage}{0.95\textwidth}
    \vskip -0.05in
    \centering
    \captionsetup{type=table}
    \caption{
      Memory complexity at different stages of our derivation in terms of $\rvx_t \in \mathbb{R}^m$, $\theta \in \mathbb{R}^n$, and the rank $R$.
      Note that \textit{all} methods have $\calO(1)$ in terms of depth.
    }
    \vskip -0.05in
    \begin{tabular}{rcccc}
      \toprule
      & Theorem~\ref{prop:1} & Proposition~\ref{prop:2} & \textbf{SNOpt} (Alg.~\ref{alg:1}) &
      first-order adjoint
       \\[1pt]
      & Eqs.~(\ref{eq:cddp-q},\ref{eq:backward-ode2})
      & Eqs.~(\ref{eq:vector-ode},\ref{eq:Quu})
      & Eqs.~(\ref{eq:q-kfac},\ref{eq:layer-kfac})
      & Eqs.~(\ref{eq:grad-node},\ref{eq:adjoint-node}) \\
      \midrule
      backward storage & $\calO((m+n)^2)$ & $\calO(Rm+Rn)$ & $\calO(Rm+2n)$ & $\calO(m+n)$ \\[3pt]
      parameter update & $\calO(n^2)$     & $\calO(n^2)$    & $\calO(2n)$    & $\calO(n)$   \\
      \bottomrule
    \end{tabular} \label{table:memory-comp}
  \end{minipage}
  \vskip -0.05in
\end{figure}

Table~\ref{table:memory-comp} summarizes the memory complexity of different computational methods
that appeared along our derivation in Section~\ref{sec:3.1} and \ref{sec:3.2}.
Despite that \textit{all} methods retain $\calO(1)$ memory as with the first-order adjoint method,
their complexity differs in terms of the state and parameter dimension.
Starting from our encouraging result in Theorem~\ref{prop:1},
which allows one to compute all derivatives with a single backward pass,
we first exploit their low-rank representation in Proposition~\ref{prop:2}.
This reduces the storage to $\calO(Rm+Rn)$ and paves a way toward adopting Kronecker factorization,
which further facilitates efficient preconditioning.
With all these, our {SNOpt} is capable of performing efficient second-order updates while enjoying similar memory complexity (up to some constant) compared to first-order adjoint methods.
Lastly, for image applications where Neural ODEs often consist of convolution layers,
we adopt convolution-based Kronecker factorization \citep{grosse2016kronecker,gao2020trace},
which effectively makes the complexity to scale w.r.t. the number of feature maps (\ie number of channels) rather than the full size of feature maps.

\subsection{Extension to Architecture Optimization} \label{sec:3.3}

\begin{wrapfigure}[12]{r}{0.32\textwidth}
  \vspace{-23pt}
  \begin{center}
    \includegraphics[height=2.35cm]{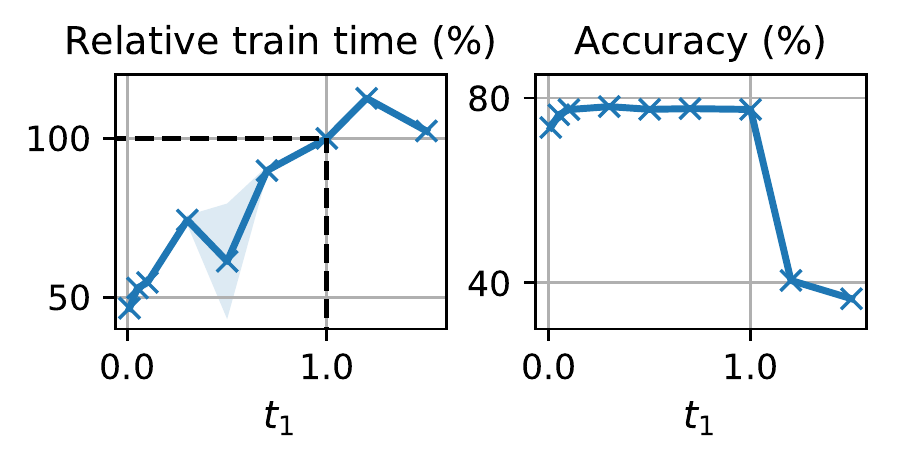}
  \end{center}
  \vskip -0.15in
  \caption{
    Training performance of CIFAR10 with Adam when using different $t_1$,
    which motivates joint optimization of $t_1$.
    Experiment setup is left in Appendix~\ref{app:4}.
  }
  \label{fig:adap-t1}
\end{wrapfigure}
Let us discuss an intriguing extension of our OCP framework to optimizing the architecture of Neural ODEs,
specifically the integration bound $t_1$.
In practice,
when problems
contain no prior information on the integration,
$[t_0, t_1]$ is typically set to
some trivial values (usually $[0,1]$) without further justification.
However,
these values can greatly affect both the performance and runtime.
Take CIFAR10 for instance (see Fig.~\ref{fig:adap-t1}),
the required training time decreases linearly as we drop $t_1$ from $1$, yet the accuracy retains mostly the same unless $t_1$ becomes too small.
Similar results also appear on MNIST (see Fig.~\ref{fig:app-adap-t1a} in Appendix~\ref{app:5}).
In other words, we may interpret the integration bound $t_1$
as an \emph{architectural parameter} that needs to be jointly optimized during training.

The aforementioned interpretation fits naturally into our OCP framework.
Specifically, we can consider the following extension of $Q$,
which introduces the terminal time $\mathrm{T}$ as a new variable:
\begin{align}
    \widetilde{Q}(t,\rvx_t,\rvu_t, \mathrm{T}) := {\widetilde{\Phi}(\mathrm{T}, \rvx(\mathrm{T}))+ \int_t^{\mathrm{T}} \ell(\tau,\rvx_\tau,\rvu_\tau) \text{ } \rd \tau },
    \label{eq:Q-T}
\end{align}
where $\widetilde{\Phi}(\mathrm{T}, \rvx(\mathrm{T}))$ explicitly imposes the penalty for longer integration time,
\eg $\widetilde{\Phi} := {\Phi}(\rvx(\mathrm{T})) + \frac{c}{2}\mathrm{T}^2$.
Following a similar procedure presented in Section~\ref{sec:3.1},
we can transform (\ref{eq:Q-T}) into its ODE form (as in (\ref{eq:Q-ode}))
then characterize its local behavior (as in (\ref{eq:cddp-q}))
along a solution path $(\hvx_t,\hvu_t,\bar{T})$.
After some tedious derivations, which are left in Appendix~\ref{app:3},
we will arrive at the update rule below,
\begin{align}
    \mathrm{T} \leftarrow \bar{T} - \eta \cdot \delta \mathrm{T}(\delta \theta), \quad \text{where}\quad
    \delta \mathrm{T}(\delta \theta)
     = [\widetilde{Q}_{\bar{T}\bar{T}}(t_0)]^{-1}\left( \widetilde{Q}_{\bar{T}}(t_0) + \widetilde{Q}_{\bar{T}\hvu}(t_0)\delta \theta \right).
    \label{eq:feedback}
\end{align}
Similar to what we have discussed in Section~\ref{sec:3.1},
one shall view $\widetilde{Q}_{\bar{T}}(t_0) \equiv \fracpartial{\calL}{\mathrm{T}}$ as the first-order derivative w.r.t. the terminal time $\mathrm{T}$. Likewise,
$\widetilde{Q}_{\bar{T}\bar{T}}(t_0) \equiv \fracpartial{^2\calL}{\mathrm{T}\partial\mathrm{T}}$, and etc.
Equation (\ref{eq:feedback}) is a second-order \emph{feedback} policy
that adjusts its updates based on the change of the parameter $\theta$.
Intuitively, it moves in the descending direction of the preconditioned gradient
(\ie $\widetilde{Q}_{\bar{T}\bar{T}}^{-1}\widetilde{Q}_{\bar{T}}$),
while accounting for the fact that \emph{$\theta$ is also progressing during training}
(via the feedback $\widetilde{Q}_{\bar{T}\hvu}\delta\theta$).
The latter is a distinct feature arising from the OCP principle.
As we will show later,
this update (\ref{eq:feedback}) leads to distinct behavior with superior convergence compared to first-order baselines \citep{massaroli2020dissecting}.

%% file: subtex/05experiment.tex
\definecolor{label1}{HTML}{99292A}
\definecolor{label2}{HTML}{D89A3C}
\definecolor{label3}{HTML}{417481}

\begin{figure}[H]
\vskip -0.15in
\begin{minipage}{0.5\textwidth}
  \centering
  \includegraphics[height=1.72cm]{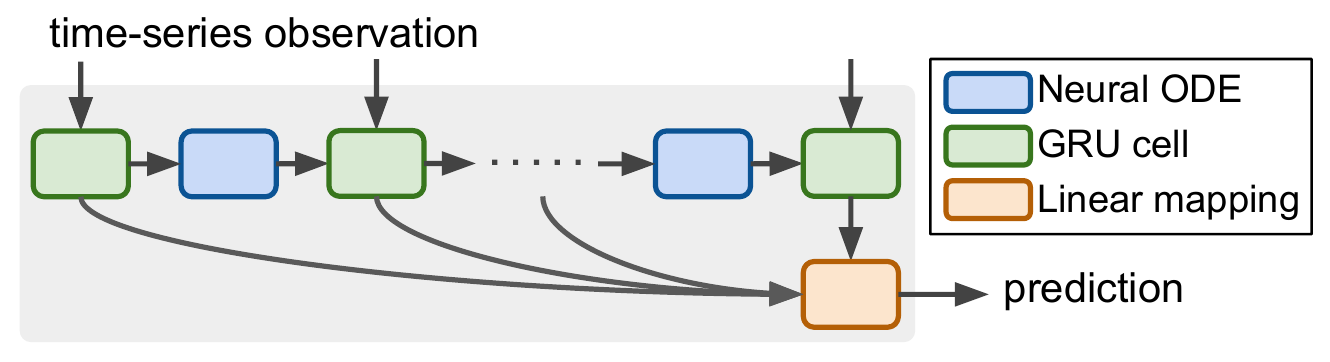}
  \vskip -0.05in
  \caption{
    Hybrid model for time-series prediction.
  }
  \label{fig:network}
\end{minipage}
\begin{minipage}{0.49\textwidth}
    \centering
    \captionsetup{type=table}
    \captionsetup{justification=centering}
    \caption{Sample size of time-series datasets \\ (input dimension, class label, series length) }
    \label{table:1}
    \vskip -0.05in
    \centering
    \begin{tabular}{ccc}
      \toprule
      SpoAD & ArtWR & CharT \\
      \midrule
      (27, 10, 93) & (19, 25, 144) & (7, 20, 187) \\
      \bottomrule
    \end{tabular}
\end{minipage}
\vskip -0.1in
\end{figure}

\textbf{Dataset.}
We select 9 datasets from 3 distinct applications where N-ODEs have been applied, including
image classification ({\color{label1}$\bullet$}), time-series prediction ({\color{label3}$\bullet$}),
and continuous normalizing flow ({\color{label2}$\bullet$}; CNF):

\vspace{-4pt}

\begin{itemize}[leftmargin=10pt]
\item[{\color{label1}$\bullet$}]
\textbf{MNIST}, \textbf{SVHN}, \textbf{CIFAR10}: MNIST consists of 28$\times$28 gray-scale images, while SVHN and CIFAR10 consist of 3$\times$32$\times$32 colour images. All 3 image datasets have 10 label classes.
\item[{\color{label3}$\bullet$}]
\textbf{SpoAD}, \textbf{ArtWR}, \textbf{CharT}: We consider UEA time series archive \citep{bagnall2018uea}.
SpokenArabicDigits (SpoAD) is a speech dataset, whereas ArticularyWordRecognition (ArtWR) and CharacterTrajectories (CharT) are motion-related datasets.
Table~\ref{table:1} details their sample sizes.
\item[{\color{label2}$\bullet$}]
\textbf{Circle}, \textbf{Gas}, \textbf{Miniboone}:
Circle is a 2-dim synthetic dataset adopted from \citet{chen2018neural}.
Gas and Miniboone are 8 and 43-dim tabular datasets commonly used in CNF \citep{grathwohl2018ffjord,onken2020ot}.
All 3 datasets transform a multivariate Gaussian to the target distributions.
\end{itemize}

\textbf{Models.}
The models for image datasets and CNF resemble standard feedforward networks, except now consisting of Neural ODEs as continuous transformation layers.
Specifically,
the models for image classification consist of convolution-based feature extraction, followed by a Neural ODE and linear mapping.
Meanwhile, the CNF models are identical to the ones in \citet{grathwohl2018ffjord}, which
consist of 1-5 Neural ODEs, depending on the size of the dataset.
As for the time-series models, we adopt the hybrid models from \cite{rubanova2019latent},
which consist of a Neural ODE for hidden state propagation,
standard recurrent cell (\eg GRU \citep{cho2014learning}) to incorporate incoming time-series observation,
and a linear prediction layer.
Figure~\ref{fig:network} illustrates this process.
We detail other configurations in Appendix~\ref{app:4}.

\textbf{ODE solver.}
We use
standard Runge-Kutta 4(5) adaptive solver (\texttt{dopri5}; \citet{dormand1980family}) implemented by the \texttt{torchdiffeq} package.
The numerical tolerance is set to 1e-6 for CNF and 1e-3 for the rest.
We fix the integration time to $[0,1]$ whenever it appears as a hyper-parameter (\eg for image and CNF datasets\footnote{
  except for Circle where we set $[t_0,t_1]{:=}[0,10]$ in order to match the original setup in \cite{chen2018neural}.
});
otherwise we adopt the problem-specific setup (\eg for time series).

\textbf{Training setup.}
We consider Adam and SGD (with momentum) as the first-order baselines since they are default training methods for most Neural-ODE applications.
As for our second-order SNOpt, we set up the time grid $\{t_j\}$ such that it collects roughly 100 samples along the backward integration to estimate the precondition matrices (see Fig.~\ref{fig:4}).
{The hyper-parameters (\eg learning rate) are tuned
for each method on each dataset,
and we detail the tuning process in Appendix~\ref{app:4}.}
We also employ practical acceleration techniques,
including the $\mathrm{semi\text{-}norm}$ \citep{kidger2020hey} for speeding up $\texttt{\markblue{ODESolve}}$, and the Jacobian-free estimator (FFJORD; \citet{grathwohl2018ffjord}) for accelerating CNF models.
The batch size is set to 256, 512, and 1000 respectively for ArtWord, CharTraj, and Gas.
The rest of the datasets use 128 as the batch size.
All experiments are conducted on a {TITAN RTX}.

\subsection{Results}

\begin{figure}[t]
\begin{minipage}{\textwidth}

    \centering
    \includegraphics[height=4.4cm]{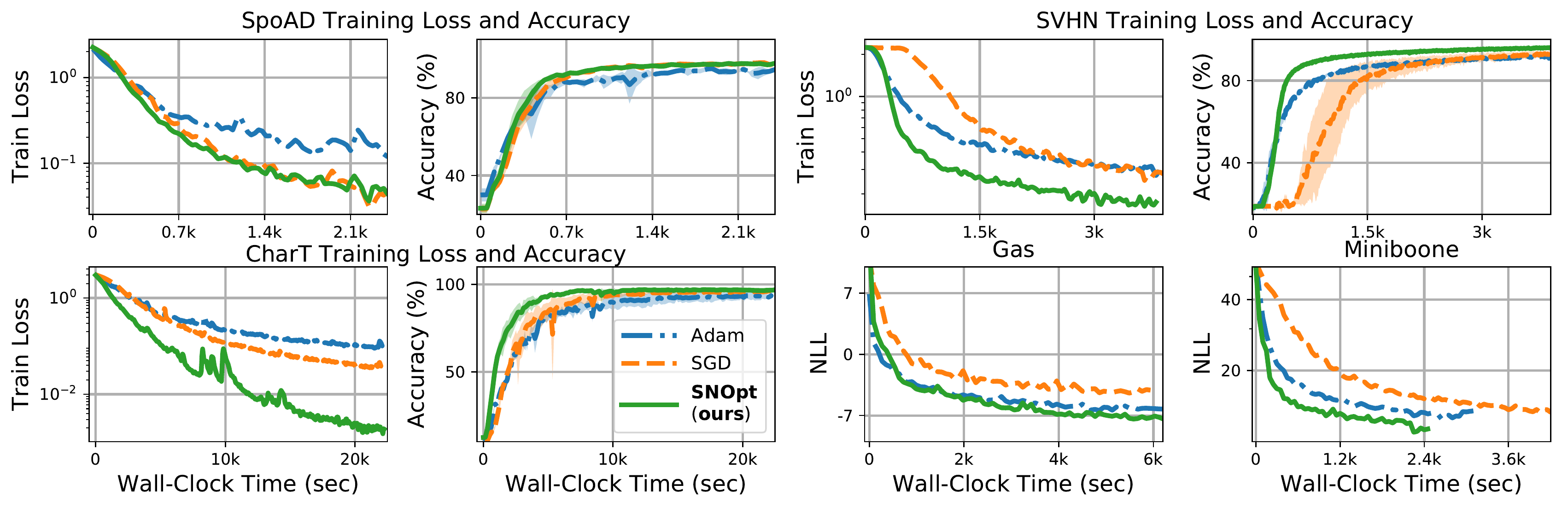}
    \vskip -0.05in
    \caption{
        Training performance in \emph{wall-clock} runtime,
        averaged over 3 trials.
        Our SNOpt achieves faster convergence against first-order baselines.
        See Fig.~\ref{fig:convergence-full} in Appendix~\ref{app:5} for MNIST and Circle.
    }
    \label{fig:convergence}
\end{minipage}
\vskip 0.15in
\begin{minipage}{\textwidth}
  \centering
  \setlength\tabcolsep{4.5pt}
  \captionsetup{type=table}
  \caption{Test-time performance: accuracies for {\color{label1}image} and {\color{label3}time-series} datasets; NLL for {\color{label2}CNF} datasets}
  \vskip -0.03in
  \centering
  \begin{tabular}{lccccccccc}
    \toprule
    & {MNIST}  & {SVHN} & {CIFAR10}
    & {SpoAD}  & {ArtWR} & {CharT}
    & Circle & {Gas}  & {Miniboone} \\
    \addlinespace[-0.25em]\arrayrulecolor{label1}
    \cmidrule[1pt](lr){2-4}\corcmidrule\arrayrulecolor{label3}%
    \cmidrule[1pt](lr){5-7}\corcmidrule\arrayrulecolor{label2}%
    \cmidrule[1pt](lr){8-10}\corcmidrule
    \addlinespace[0.4em]\arrayrulecolor{black}
    \midrule
    Adam          & 98.83          &  91.92          & 77.41          & 94.64          & 84.14          & 93.29
                  & 0.90           & -6.42          &  13.10 \\[2pt]
    SGD           & 98.68          &  93.34          & 76.42          & \textbf{97.70} & 85.82          & 95.93
                  & 0.94           & -4.58          &  13.75 \\[1pt]
    \midrule
    \textbf{SNOpt}
                  & \textbf{98.99} &  \textbf{95.77} & \textbf{79.11} & 97.41          & \textbf{90.23} & \textbf{96.63}
                  &\textbf{0.86}  & \textbf{-7.55} &  \textbf{12.50} \\
    \bottomrule
  \end{tabular} \label{table:accu}
\end{minipage}
\vskip 0.15in
\begin{minipage}{0.37\textwidth}
    \vskip 0.05in
    \centering
    \includegraphics[height=3.5cm]{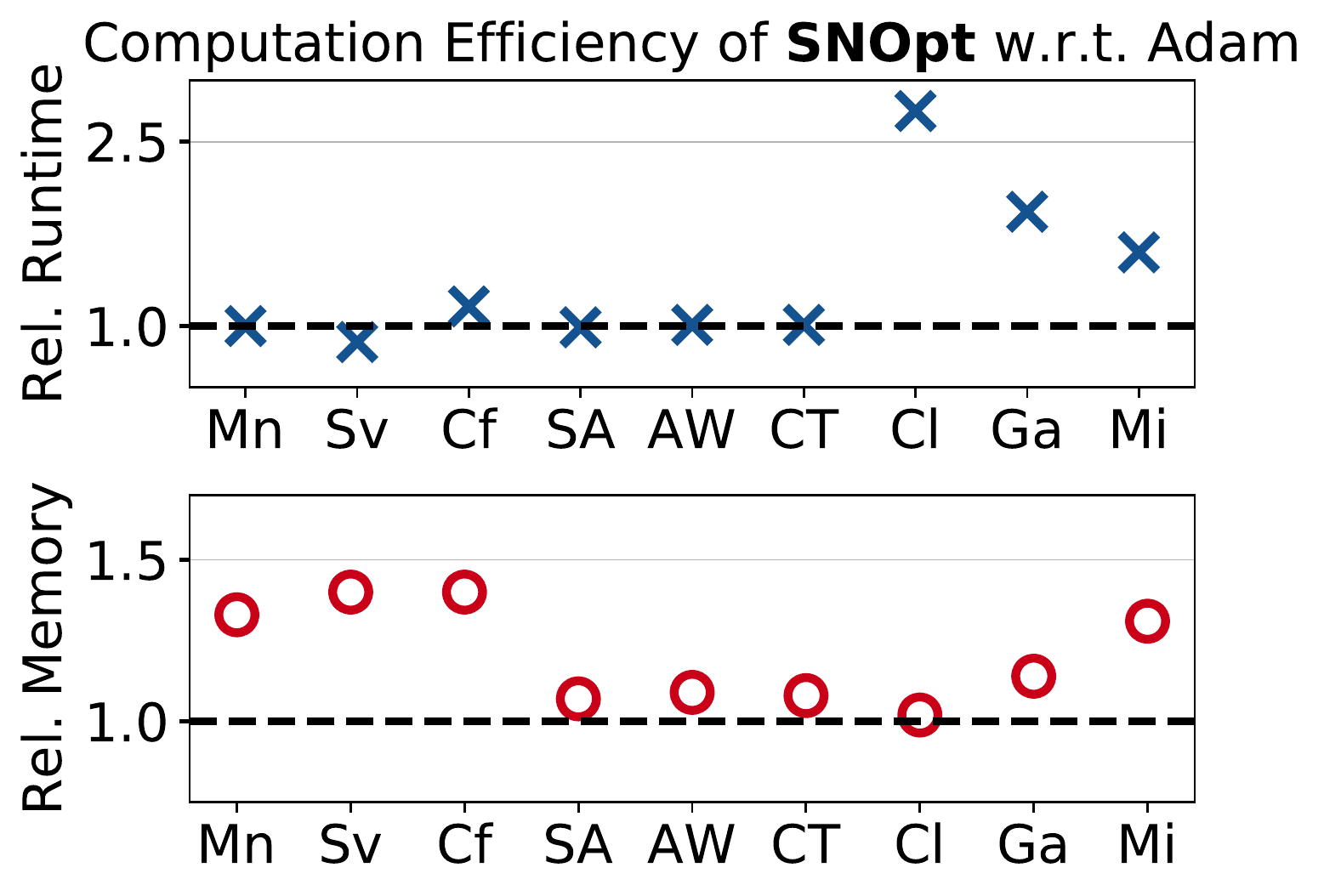}
    \caption{
        Relative runtime and memory of our SNOpt compared to Adam (denoted by the dashed black lines) on all 9 datasets, where `Mn' is the shorthand for MNIST, and \textit{etc}.
    }
    \label{fig:complexity}
\end{minipage}
\hfill
\begin{minipage}{0.6\textwidth}
    \centering
    \includegraphics[height=3.8cm]{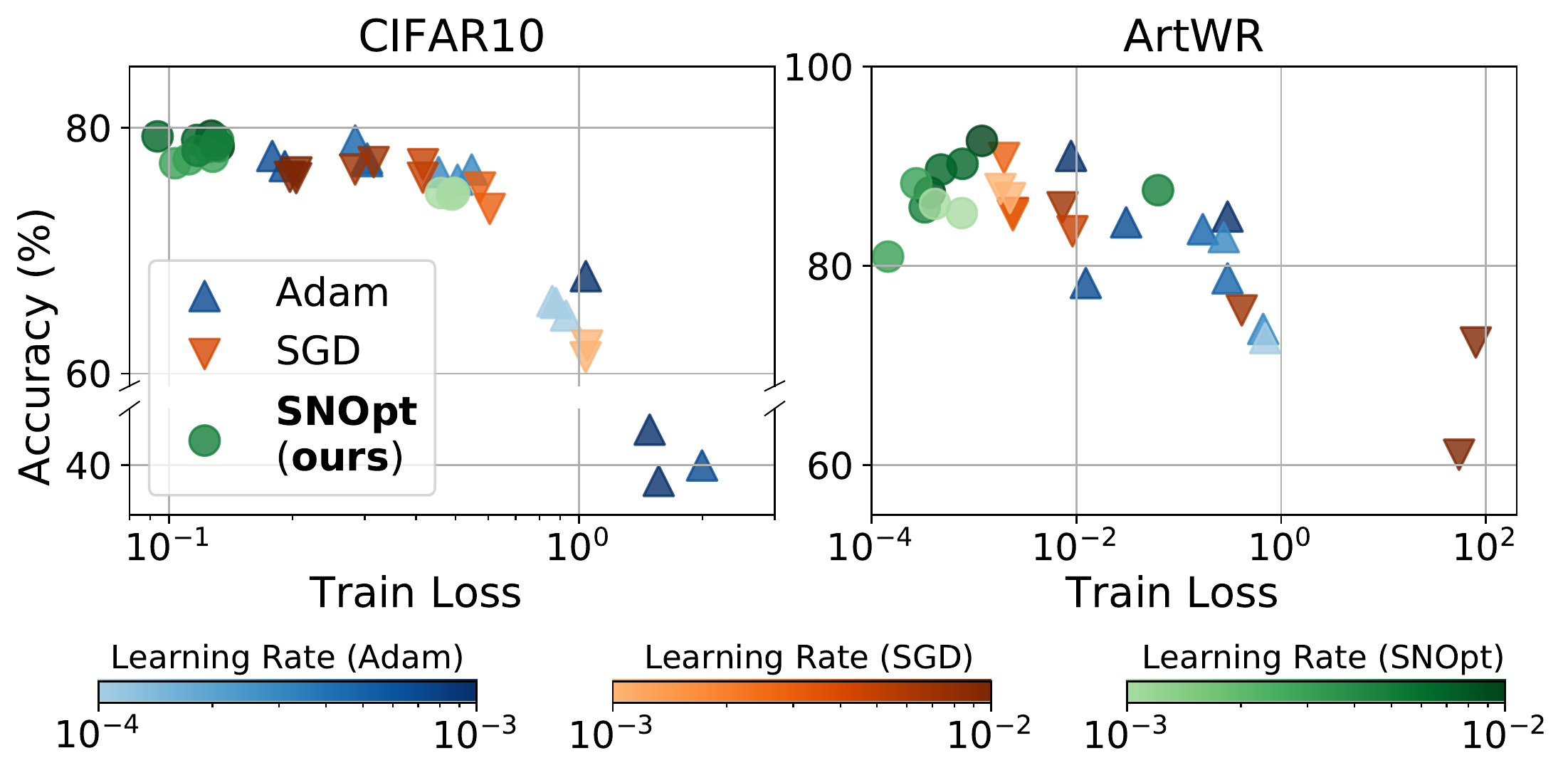}
    \vskip -0.05in
    \caption{
      Sensitivity analysis where
      each sample represents a training result using different optimizer and learning rate
      (annotated by different symbol and color).
      Our SNOpt achieves higher accuracies and is insensitive to hyper-parameter changes.
      Note that x-axes are in $\log$ scale.
    }
    \label{fig:point-cloud}
\end{minipage}
\vskip -0.15in
\end{figure}

\textbf{Convergence and computation efficiency.}
Figures~\ref{fig:1} and \ref{fig:convergence}
report the training curves of each method measured by wall-clock time.
It is obvious that our SNOpt admits a superior
convergence rate compared to the first-order baselines,
and in many cases exceeds their performances by a large margin.
In Fig.~\ref{fig:complexity},
we report the computation efficiency of our SNOpt compared to Adam on each dataset,
and leave their numerical values in Appendix~\ref{app:4} (Table~\ref{table:runtime} and \ref{table:memory}).
For image and time-series datasets (\ie Mn{\texttildelow}CT),
our SNOpt
runs nearly as fast as first-order methods.
This is made possible
through a rigorous OCP analysis in Section~\ref{sec:3},
where we showed that second-order matrices %
can be constructed along with the \emph{same} backward integration when we compute the gradient.
Hence, only a minimal overhead is introduced.
As for CNF, which propagates the probability density additional to the vanilla state dynamics,
our SNOpt is roughly 1.5 to 2.5 times slower,
yet it still converges faster in the overall wall-clock time (see Fig.~\ref{fig:convergence}).
On the other hand,
the use of second-order matrices
increases the memory consumption of SNOpt by 10-40\%,
depending on the model and dataset.
However,
the actual increase in memory (less than 1GB for all datasets; see Table~\ref{table:memory})
remains affordable
on standard GPU machines.
More importantly, our SNOpt retains the $\calO(1)$ memory throughout training.

\textbf{Test-time performance and hyper-parameter sensitivity.}
Table~\ref{table:accu} reports the test-time performance,
including the accuracies (\%) for image and time-series classification,
and the negative log-likelihood (NLL) for CNF.
On most datasets, our method achieves competitive results against standard baselines.
In practice, we also find that using the preconditioned updates
greatly reduce the sensitivity to hyper-parameters (\eg learning rate).
This is demonstrated in Fig.~\ref{fig:point-cloud},
where we sample distinct learning rates from a proper interval for each method (shown with different color bars)
and record their training results after convergence.
It is clear that
our method not only converges to higher accuracies with lower losses,
these values are also more concentrated on the plots.
In other words,
our method achieves better convergence in a more consistent manner across different hyper-parameters.

\begin{figure}[t]
\vskip -0.15in
\centering
\begin{minipage}{0.43\textwidth}
    \setlength\tabcolsep{3pt}
    \centering
    \captionsetup{type=table}
    \caption{
        Performance of jointly optimizing the integration bound $t_1$ on CIFAR10
    }
    \vskip -0.07in
    \centering
    \begin{tabular}{lcc}
      \toprule
      Method & \specialcell{Train time (\%) \\ w.r.t. $t_1{=}1.0$ }
      & \specialcell{Accuracy \\ (\%)}
         \\
      \midrule
      ASM baseline
      &   96   & 76.61    \\
      \textbf{SNOpt (ours)}  & \textbf{81} & \textbf{77.82} \\
      \bottomrule
    \end{tabular} \label{table:adap-t1}
    \begin{center}
    \captionsetup{type=figure}
    \includegraphics[height=2cm]{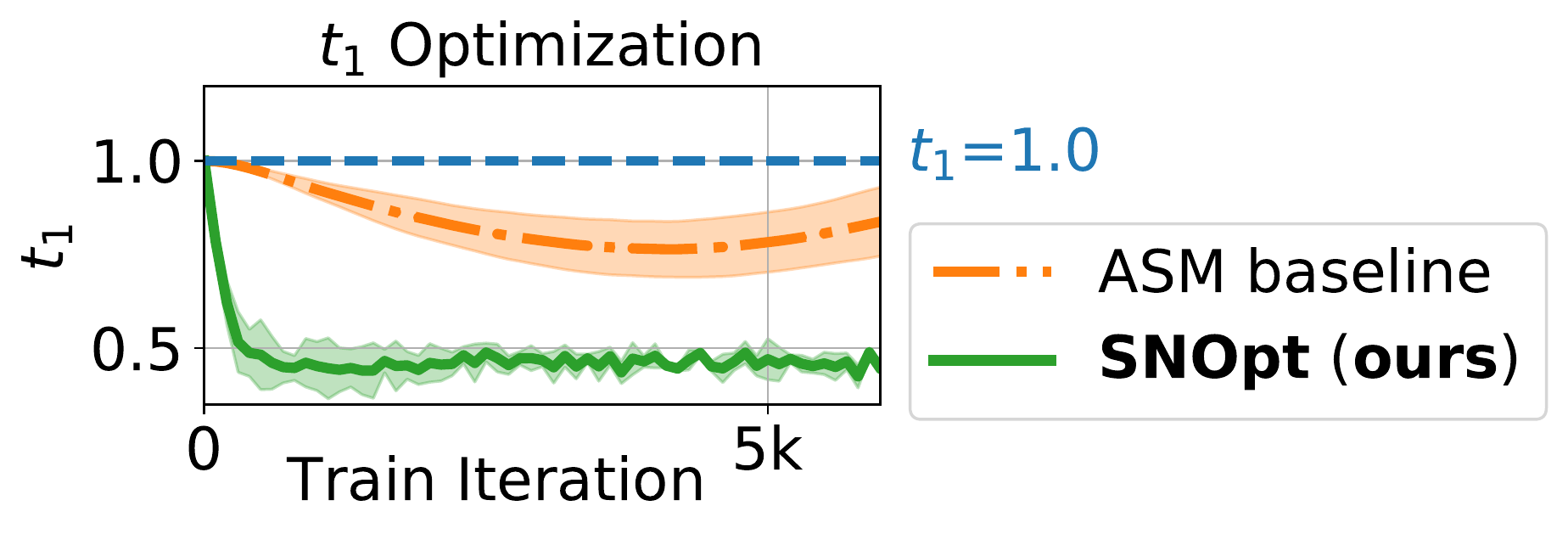}
    \caption{
      Dynamics of $t_1$ over CIFAR10 training using different methods.
    }
    \label{fig:adap-t1b}
    \end{center}
\end{minipage}
\hfill
\begin{minipage}{0.55\textwidth}
    \setlength\tabcolsep{3pt}
    \centering
    \captionsetup{type=table}
    \caption{
        Measure of implicit regularization on SVHN
    }
    \vskip -0.07in
    \centering
    \begin{tabular}{lcc}
      \toprule
      & \# of function & Regularization \\
      & evaluation (NFE) & ($ \int \norm{\nabla_\rvx F}^2 + \int \norm{F}^2$) \\
      \midrule
      Adam
      &   42.1   & 323.88    \\
      \textbf{SNOpt}  & \textbf{32.6} & \textbf{199.1} \\
      \bottomrule
    \end{tabular} \label{table:reg}
    \begin{center}
    \captionsetup{type=figure}
    \includegraphics[height=2cm]{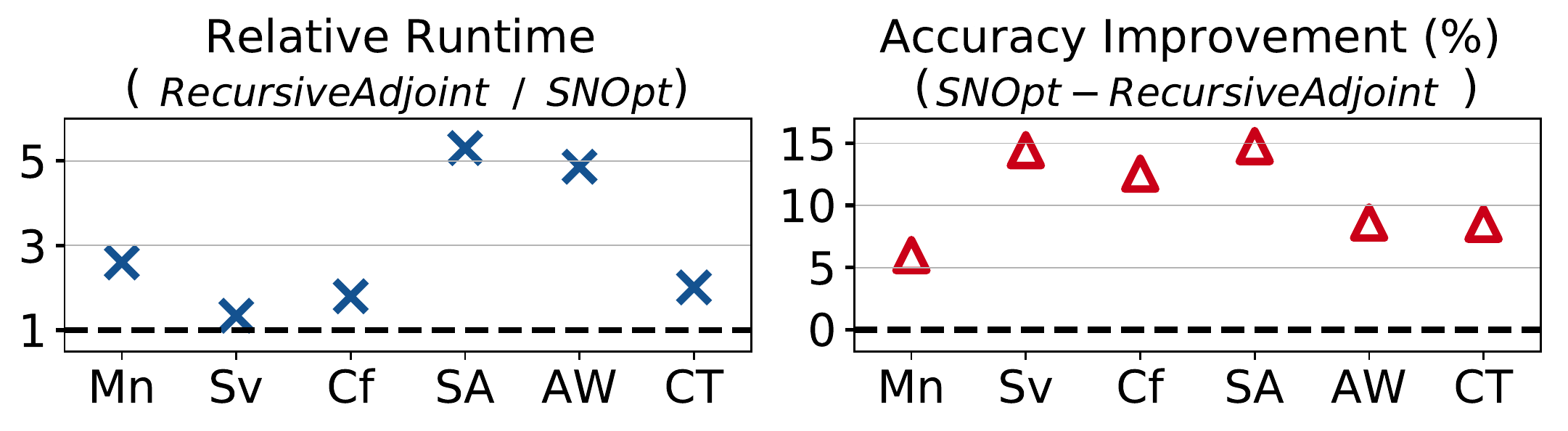}
    \caption{
      Comparison between SNOpt and second-order recursive adjoint.
      SNOpt is at least 2 times faster and improves the accuracies of baselines by 5-15\%.
    }
    \label{fig:recur-adj}
    \end{center}
\end{minipage}
\vskip -0.1in
\end{figure}

\textbf{Joint optimization of the integration bound $t_1$.}
Table~\ref{table:adap-t1} and Fig.~\ref{fig:adap-t1b}
report the performance of optimizing $t_1$ along with its convergence dynamics.
Specifically, we compare our second-order feedback policy (\ref{eq:feedback}) derived in Section~\ref{sec:3.3}
to the first-order ASM baseline proposed in \citet{massaroli2020dissecting}.
It is clear that
our OCP-theoretic method
leads to substantially faster convergence,
and the optimized $t_1$ stably hovers around $0.5$ without deviation (as appeared for the baseline).
This drops the training time by nearly 20\% compared to the vanilla training, where we fix $t_1$ to $1.0$,
yet without sacrificing the test-time accuracy.
A similar experiment for MNIST (see Fig.~\ref{fig:app-adap-t1b} in Appendix~\ref{app:5})
shows a consistent result.
We highlight
these improvements as the benefit gained from introducing the well-established OCP principle
to these emerging deep continuous-time models.

\textbf{Comparison with recursive adjoint.}
Finally,
Fig.~\ref{fig:recur-adj} reports the comparison between our SNOpt and the recursive adjoint baseline (see Section~\ref{sec:2} and Table~\ref{table:adj2-error}).
It is clear that our method outperforms this second-order baseline by a large margin
in both runtime efficiency and test-time performance.
Note that we omit the comparison on CNF datasets since the recursive adjoint simply fails to converge.

\textbf{Remark} (Implicit regularization)\textbf{.}
  In some cases (\eg SVHN in Fig.~\ref{fig:complexity}),
  our method may run slightly faster
  than first-order methods.
  This is a distinct phenomenon arising exclusively from training these continuous-time models.
  Since their forward and backward passes involve solving \emph{parameterized} ODEs (see Fig.~\ref{fig:2}),
  the computation graphs are \emph{parameter-dependent}; hence adaptive throughout training.
  In this vein, we conjecture that
  the preconditioned updates in these cases may have guided the parameter to regions that
  are numerically stabler (hence faster) for integration.\footnote{
    In Appendix~\ref{app:4}, we provide some theoretical discussions (see Corollary~\ref{coro:4}) in this regard.
    \label{footnote:4}
  }
  With this in mind, we report in Table~\ref{table:reg}
  the value of Jacobian, $ \int \norm{\nabla_\rvx F}^2$, and Kinetic, $\int \norm{F}^2$, regularization \citep{finlay2020train} in SVHN training.
  Interestingly, the parameter found by our SNOpt indeed has a substantially lower
  value (hence stronger regularization and better-conditioned ODE dynamics) compared to the one found by Adam. This provides a plausible explanation of the reduction in the NFE when using our method, yet without hindering the test-time performance (see Table~\ref{table:accu}).

%% file: subtex/07appendix.tex
Here, we review the OCP perspective of training discrete DNNs and discuss
how the continuous-time OCP can be connected to the training process of Neural ODEs.
For a complete treatment, we refer readers to \eg
\citet{weinan2017proposal,li2017maximum,han2018mean,liu2019deep,liu2021differential},
and their references therein.

Abuse the notation and let the layer propagation rule in standard {feedforward} DNNs with depth $T$ be
\begin{align}
    \vz_{t+1} =& f(\vz_{t}, \vu_{t}),
    \quad t \in \{0,1,\cdots,T\}.
    \label{eq:layer-prop}
\end{align}
Here,
$\vz_t$ and $\vu_t$
represent the (vectorized) hidden state and parameter of layer $t$.
For instance, consider the propagation of a fully-connected layer, \ie
$\vz_{t+1} = \sigma(\mW_t \vz_t { +} \vb_t)$,
where $\mW_t$, $\vb_t$, and $\sigma(\cdot)$ are respectively the weight, bias, and nonlinear activation function.
Then,
(\ref{eq:layer-prop}) treats
$\vu_t \coloneqq \vectorize([\mW_t, \vb_t])$ as the vectorized parameter
and $f$ as the composition of $\sigma(\cdot)$ and the affine transformation
 (Do not confuse with Fig.~\ref{fig:3} which denotes $f$ as the affine transformation).

The OCP perspective notices that
(\ref{eq:layer-prop}) can also be interpreted as a discrete-time dynamical system that propagates the state $\vz_t$ with the {control} variable $\vu_t$.
In this vein,
computing the forward pass of a DNN can be seen as propagating a nonlinear dynamical system from time $t=0$ to $T$.
Furthermore, the training process, \ie finding optimal parameters $\{\vu_t: \forall t \}$ for all layers,
can be seen as a discrete-time Optimal Control Programming (OCP),
which searches for an optimal {control sequence} $\{\vu_t:  \forall t \}$ that minimizes some objective.

In the case of Neural ODEs, the discrete-time layer propagation rule in (\ref{eq:layer-prop}) is replaced with the ODE in (\ref{eq:node}). However, as we have shown in Section~\ref{sec:3.1}, the interpretation between the trainable parameter $\theta$ and control variable (hence the connection between the training process and OCP) remains valid.
In fact, consider the vanilla form of continuous-time OCP,
\begin{equation}
\begin{split}
\min_{\rvu(t): t\in [t_0, t_1]} \br{\Phi(\rvx_{t_1}) + \intT \ell(t,\rvx_t, \rvu_t) \dt }, \quad
\dot{\rvx}_{t} = F(t, \rvx_t, \rvu_t), \quad & \rvx_{t_0} = \vx_{t_0},
\label{eq:ct-ocp-app}
\end{split}
\end{equation}
which resembles the one we used in (\ref{eq:ct-ocp}) except considering a time-varying control process $\rvu(t)$.
The necessary condition to the programming (\ref{eq:ct-ocp-app}) can be characterized by the
celebrated Pontryagin's maximum principle \citep{pontryagin1962mathematical}.
\begin{theorem}[Pontryagin's maximum principle] \label{thm:pmp}
    Let $\rvu^*_t \equiv \rvu^*(t)$ be a solution that achieved the minimum of (\ref{eq:ct-ocp-app}). Then, there exists
    continuous processes, $\rvx^*_t$ and $\rva^*_t$, such that
    \begin{subequations} \label{eq:pmp}
    \begin{alignat}{3}
    &\dot{\rvx}_{t}^{*}= \nabla_{\rva} H\left(t, \rvx_{t}^{*}, \rva_{t}^{*}, \rvu_{t}^{*}\right)
    \qquad &&\rvx_{0}^{*}=\rvx_{0}
    \comma \label{eq:pmp-forward} \\
      &\dot{\rva}_{t}^{*}=-\nabla_{\rvx} H\left(t, \rvx_{t}^{*}, \rva_{t}^{*}, \rvu_{t}^{*}\right),
    \qquad &&\rva_{t_1}^{*}= \nabla_{\rvx} \Phi\left(\rvx_{t_1}^{*} \right)
    \comma \label{eq:pmp-backward}\\
    & H\left(t, \rvx_{t}^{*}, \rva_{t}^{*}, \rvu_t^{*}\right)
        \leq
            H\left(t, \rvx_{t}^{*}, \rva_{t}^{*}, \rvu_t \right),
    \qquad &&\forall \rvu_t \in \cspace, \quad t \in [t_0, t_1]
    \comma \label{eq:pmp-max-h}
    \end{alignat}
    \end{subequations}
    where the Hamiltonian function
    is defined as
    \begin{align*}
    H\left(t, \rvx_t, \rva_t, \rvu_t \right) \coloneqq \rva_t \cdot F(t, \rvx_t, \rvu_t) + \ell(t, \rvx_t, \rvu_t).
    \end{align*}
\end{theorem}
It can be readily verified that ({\ref{eq:pmp-backward}}) gives the same backward ODE in (\ref{eq:adjoint-node}).
In other words, the Adjoint Sensitivity Method used for deriving (\ref{eq:grad-node},~\ref{eq:adjoint-node})
is a direct consequence arising from the OCP optimization theory.
In this work, we provide a full treatment of continuous-time OCP theory and show that
it opens up new algorithmic opportunities to higher-order training methods for Neural ODEs.

%% file: subtex/08appendix2.tex
\textbf{Proof of Theorem~\ref{prop:1}.}
Rewrite the backward ODE of the accumulated loss $Q$ in (\ref{eq:Q-ode}) below
\begin{align*}
{0 = {\ell(t,\rvx_t,\rvu_t) + \fracdiff{Q(t,\rvx_t,\rvu_t)}{t}}}, \quad
Q({t_1},\rvx_{t_1}) = \Phi(\rvx_{t_1}).
\end{align*}
Given a solution path $(\hvx_t, \hvu_t)$ of the ODEs in (\ref{eq:ct-ocp}),
define the {differential} state and control variables $(\drvx_t,\drvu_t)$ by
\begin{align*}
    \drvx_t := \rvx_t - \hvx_t \quad \text{and} \quad \drvu_t := \rvu_t - \hvu_t.
\end{align*}
We first perform second-order expansions for $\ell$ and $Q$ along the solution path,
which are given by
\begin{subequations}
\begin{alignat}{1}
\ell
\approx&\text{ }
\ell(t, \hvx_t,\hvu_t) +
\Lx^\T \drvxt + \Lu^\T \drvut +
\frac{1}{2} {\bvec{\drvxt \\ \drvut}}^\T \bmat{\Lxx}{\Lxu}{\Lux}{\Luu} \bvec{\drvxt \\ \drvut}, \label{eq:l-expansion} \\
Q
\approx&\text{ }
Q(t, \hvx_t,\hvu_t) +
\Qx^\T \drvxt + \Qu^\T \drvut +
\frac{1}{2} {\bvec{\drvxt \\ \drvut}}^\T \bmat{\Qxx}{\Qxu}{\Qux}{\Quu} \bvec{\drvxt \\ \drvut}, \label{eq:q-expansion}
\end{alignat} \label{eq:l-q-expansion}
\end{subequations}
where all derivatives, \ie $\Lx, \Lu, \Qxx, \Quu$, and etc, are time-varying.
We can thereby obtain the time derivative of the second-order approximated $Q$ in (\ref{eq:q-expansion}) following standard ordinary calculus.
\begin{equation}
\begin{split}
\fracdiff{Q}{t}
\approx& \fracdiff{Q(t, \hvx_t,\hvu_t)}{t}
+ \pr{{\fracdiff{\Qx}{t}}^\T \drvxt + \Qx^\T \fracdiff{\drvxt}{t} }
+ \pr{{\fracdiff{\Qu}{t}}^\T \drvut + \Qu^\T \fracdiff{\drvut}{t} } \\
&\quad
+ \frac{1}{2} \pr{{\drvxt^\T \fracdiff{\Qxx}{t}} \drvxt + {\fracdiff{\drvxt}{t}}^\T \Qxx \drvxt + \drvxt^\T \Qxx \fracdiff{\drvxt}{t} } \\
&\quad
+ \frac{1}{2} \pr{{\drvut^\T \fracdiff{\Quu}{t}} \drvut + {\fracdiff{\drvut}{t}}^\T \Quu \drvut + \drvut^\T \Quu \fracdiff{\drvut}{t} } \\
&\quad
+ \frac{1}{2} \pr{{\drvxt^\T \fracdiff{\Qxu}{t}} \drvut + {\fracdiff{\drvxt}{t}}^\T \Qxu \drvut + \drvxt^\T \Qxu \fracdiff{\drvut}{t} } \\
&\quad
+ \frac{1}{2} \pr{{\drvut^\T \fracdiff{\Qux}{t}} \drvxt + {\fracdiff{\drvut}{t}}^\T \Qux \drvxt + \drvut^\T \Qux \fracdiff{\drvxt}{t} }.
\label{eq:Q-2nd-expand}
\end{split}
\end{equation}
Next, we need to compute $\fracdiff{\drvxt}{t}$ and $\fracdiff{\drvut}{t}$, \ie
the dynamics of the differential state and control. This can be achieved by
linearizing the ODE dynamics along $(\hvx_t,\hvu_t)$.
\begin{align*}
\fracdiff{}{t}(\hvx_t+\drvxt) = {F(t, \hvx_t,\hvu_t)+\Fx(t)^\T \drvxt + \Fu(t)^\T \drvut}
\text{ }\Rightarrow&\text{ } \fracdiff{\drvxt}{t} = {\Fx(t)^\T \drvxt + \Fu(t)^\T \drvut}, \\
\fracdiff{}{t}(\hvu_t+\drvut) = \mathbf{0}
\text{ }\Rightarrow&\text{ } \fracdiff{\drvut}{t} = \mathbf{0},
\numberthis \label{eq:dxt-dut}
\end{align*}
since $\fracdiff{\hvx_t}{t} =F(t,\hvx_t,\hvu_t)$.
Finally,
substituting (\ref{eq:l-expansion}) and (\ref{eq:Q-2nd-expand}) back to (\ref{eq:Q-ode}) and replacing all
$(\fracdiff{\drvxt}{t}, \fracdiff{\drvut}{t})$ with (\ref{eq:dxt-dut})
yield the following set of backward ODEs.
\begin{subequations}
\begin{alignat*}{2}
- \fracdiff{\Qx}{t} &= \Lx + \FxT\Qx, \quad
&&\text{ }\text{ }- \fracdiff{\Qu}{t} = \Lu + \FuT\Qx,  \\
- \fracdiff{\Qxx}{t} &= \Lxx + \FxT\Qxx+\Qxx\Fx, \quad
&&- \fracdiff{\Qxu}{t} = \Lxu + \Qxx\Fu + \FxT\Qxu, \\
- \fracdiff{\Quu}{t} &= \Luu + \FuT\Qxu + \Qux\Fu, \quad
&&- \fracdiff{\Qux}{t} = \Lux + \FuT\Qxx + \Qux\Fx.
\end{alignat*}
\end{subequations}
\hfill $\qedsymbol$

\begin{remark}[Relation to continuous-time OCP algorithm]\normalfont
The proof of Theorem~\ref{prop:1} resembles standard derivation of
continuous-time Differential Dynamic Programming (DDP),
a second-order OCP method that has shown great successes in modern autonomous systems \citep{tassa2014control}.
However, our derivation was modified accordingly to account for the particular OCP proposed in (\ref{eq:ct-ocp}), which concerns only the initial condition of the time-invariant control.
As this equivalently leaves out the ``dynamic'' aspect of DDP, we shorthand our methodology by \emph{Differential Programming}.
\end{remark}

\begin{remark}[Computing higher-order derivatives]\normalfont\label{r:higher-order}
The proof of Theorem~\ref{prop:1} can be summarized by
\begin{enumerate}[leftmargin=40pt]
\item[\textit{Step 1}.] Expand $Q$ and $\ell$ up to second-order, \ie (\ref{eq:l-q-expansion}).

\item[\textit{Step 2}.] Derive the dynamics of differential variables. In our case, we consider the linear ODE presented in (\ref{eq:dxt-dut}).
\item[\textit{Step 3}.] Substitute the approximations from Step 1 and 2 back to (\ref{eq:Q-ode}), expand all terms using ordinary calculus (\ref{eq:Q-2nd-expand}), then collect the dynamics of each derivative.
\end{enumerate}
For higher-order derivatives, we simply need to consider a higher-order expansion of $Q$ and $\ell$ in Step 1
(see \eg \citet{almubarak2019infinite} and their reference therein).
It is also possible to consider higher-order expression of the linear differential ODEs in Step 2,
which may further improve the convergence at the cost of extra overhead \citep{theodorou2010stochastic}.
\end{remark}

\begin{remark}[Complexity of Remark~\ref{r:higher-order}]\normalfont
Let $k$ be the optimization order.
Development of higher-order ($k\ge$3) optimization based on Theorem~\ref{prop:1} certainly has few computational obstacles, just like what we have identified and resolved in the case of $k=$2 (see Section~\ref{sec:3.2}). In terms of memory,
while the number of backward ODEs suggested by Theorem~\ref{prop:1} can grow exponentially w.r.t. $k$,
\citet{kelly2020learning} has developed an efficient truncated method that reduces the number to $\calO(k^2)$ or $\calO(k \log k)$. In terms of runtime, analogous to the Kronecker approximation that we use to factorize second-order matrices, \citet{gupta2018shampoo} provided an extension to generic higher-order tensor programming. Hence, it may still be plausible to avoid impractical training.
\end{remark}

\allowdisplaybreaks

\textbf{Proof of Proposition~\ref{prop:2}.}
We will proceed the proof by induction.
Recall that when $\ell$ degenerates, the matrix ODEs presented in (\ref{eq:cddp-qb}, \ref{eq:cddp-qc}) from Theorem~\ref{prop:1} take the form,
\begin{subequations}
\begin{alignat}{2}
- \fracdiff{\Qxx}{t} &= \FxT\Qxx+\Qxx\Fx, \qquad && \Qxx(t_1) = \Phi_{\hvx\hvx},  \\
- \fracdiff{\Quu}{t} &= \Fu^\T\Qxu + \Qux\Fu , \qquad && \Quu(t_1) = \mathbf{0}, \\
- \fracdiff{\Qxu}{t} &= \Qxx\Fu + \Fx^\T\Qxu , \qquad && \Qxu(t_1) = \mathbf{0},
\end{alignat} \label{eq:cddp-q-app}
\end{subequations}
where we leave out the ODE of $\Qux$ since $\Qux(t) = Q_{\hvx \hvu}^\T(t)$ for all $t\in[t_0,t_1]$.

From (\ref{eq:cddp-q-app}), it is obvious that the decomposition given in Proposition~\ref{prop:2} holds at the terminal stage $t_1$.
Now, suppose it also holds at $t \in (t_0, t_1)$, then the backward dynamics of second-order matrices at this specific time step $t$,
take $\fracdiff{\Qxx(t)}{t}$ for instance, become
\begin{align*}
- \fracdiff{\Qxx}{t}
&= \FxT\Qxx+\Qxx\Fx \\
&= \FxT \left( \sum_{i=1}^R \rvq_i \otimes \rvq_i \right) + \left( \sum_{i=1}^R \rvq_i \otimes \rvq_i \right) \Fx \\
&= \sum_{i=1}^R \left[ \left(\FxT\rvq_i\right) \otimes \rvq_i + \rvq_i \otimes \left(\FxT\rvq_i\right) \right],
\numberthis \label{eq:low-rank-Qxx}
\end{align*}
where $\rvq_i \equiv \rvq_i(t)$ for brevity.
On the other hand, the LHS of (\ref{eq:low-rank-Qxx}) can be expanded as
\begin{align}
- \fracdiff{\Qxx}{t}
= - \fracdiff{}{t}\left( \sum_{i=1}^R \rvq_i \otimes \rvq_i \right)
= - \sum_{i=1}^R \left[ \fracdiff{\rvq_i}{t} \otimes \rvq_i + \rvq_i \otimes \fracdiff{\rvq_i}{t}  \right],
\label{eq:low-rank-Qxx2}
\end{align}
which follows by standard ordinary calculus.
Equating (\ref{eq:low-rank-Qxx}) and (\ref{eq:low-rank-Qxx2}) implies that following relation should hold at time $t$,
\begin{align*}
- \fracdiff{\rvq_i}{t} = \FxT\rvq_i,
\end{align*}
which yields the first ODE appeared in (\ref{eq:vector-ode}).
Similarly, we can repeat the same process (\ref{eq:low-rank-Qxx}, \ref{eq:low-rank-Qxx2}) for the matrices $\Qxu$ and $\Quu$.
This will give us
\begin{align*}
- \fracdiff{\Quu}{t} =& \Fu^\T\Qxu + \Qux\Fu \\
\Rightarrow&
-\sum_{i=1}^R \left[ \fracdiff{\rvp_i}{t} \otimes \rvp_i + \rvp_i \otimes \fracdiff{\rvp_i}{t}  \right]
=
\sum_{i=1}^R \left[ \left(\FuT\rvq_i\right) \otimes \rvp_i + \rvp_i \otimes \left(\FuT\rvq_i\right) \right]
\\
- \fracdiff{\Qxu}{t} =& \Qxx\Fu + \Fx^\T\Qxu \\
\Rightarrow&
-\sum_{i=1}^R \left[ \fracdiff{\rvq_i}{t} \otimes \rvp_i + \rvq_i \otimes \fracdiff{\rvp_i}{t}  \right]
=
\sum_{i=1}^R \left[ \left(\FxT\rvq_i\right) \otimes \rvp_i + \rvq_i \otimes \left(\FuT\rvq_i\right) \right],
\end{align*}
which implies that following relation should also hold at time $t$,
\begin{align*}
- \fracdiff{\rvp_i}{t} = \FuT\rvq_i.
\end{align*}
Hence, we conclude the proof.
\hfill $\qedsymbol$

\textbf{Derivation and approximation in (\ref{eq:q-kfac},~\ref{eq:layer-kfac}).}
We first recall two formulas related to the Kronecker product that will be shown useful in deriving (\ref{eq:q-kfac},~\ref{eq:layer-kfac}).
\begin{align}
    (\mA \otimes \mB)(\mC \otimes \mD)^\T &= \mA\mC^\T \otimes \mB\mD^\T,
    \label{eq:kfac-formula1}
    \\
      (\mA \otimes \mB)^\Inv \vectorize (\mW)
    &= \vectorize (\mB^\Inv \mW \mA^{-\transpose}),
    \label{eq:kfac-formula2}
\end{align}
where $\mW \in \mathbb{R}^{l\times p} $, $\mA, \mC \in \mathbb{R}^{p \times p}$, and $\mB, \mD \in \mathbb{R}^{l\times l}$.
Further, $\mA, \mB$ are invertible.

Now, we provide a step-by-step derivation of (\ref{eq:q-kfac}).
For brevity, we will denote $ \rvg^n_i \equiv {\fracpartial{F}{\rvh^n}}^\T \rvq_i$.
\begin{align*}
\calL_{\theta^n\theta^n}
\equiv
Q_{\hvu^n \hvu^n}(t_0)
&=
\sum_{i=1}^R
\left( \rintT \left( \textstyle \rvz^n \otimes \rvg^n_i \right) \dt \right)
\left( \rintT \left( \textstyle
\rvz^n \otimes \rvg^n_i \right) \dt \right)^\T \\
&\approx
\sum_{i=1}^R
\rintT \left( \textstyle \rvz^n \otimes \rvg^n_i \right)
\left( \textstyle
\rvz^n \otimes \rvg^n_i \right)^\T \dt
\\
&=
\sum_{i=1}^R
\rintT
\left( \textstyle \rvz^n {\rvz^n}^\T \right)
\otimes
\left( \textstyle \rvg^n_i {\rvg^n_i}^\T \right)
\dt
&& \text{by (\ref{eq:kfac-formula1})}
\\
&\approx
\sum_{i=1}^R
\rintT \left( \textstyle \rvz^n {\rvz^n}^\T \right) \dt
\otimes
\rintT \left( \textstyle \rvg^n_i {\rvg^n_i}^\T \right) \dt
\\
&=
\rintT \left( \textstyle \rvz^n \otimes \rvz^n \right) \dt
\otimes
\rintT \sum_{i=1}^R \rvg^n_i \otimes \rvg^n_i \dt.
&& \text{by Fubini's Theorem}
\end{align*}
There are two approximations in the above derivation.
The first one
assumes that the contributions of the quantity ``$\rvz^n(t) \otimes \rvg^n_i(t)$''
are uncorrelated across time,
whereas the second one assumes $\rvz^n$ and $\rvg^n_i$ are pair-wise independents.
We stress that both are widely adopted assumptions for deriving practical Kronecker-based methods \citep{grosse2016kronecker,martens2018kronecker}.
While the first assumption can be rather strong, the second approximation
has been verified in some empirical study \citep{wu2020dissecting} and
can be made exact under certain conditions \citep{martens2015optimizing}.
Finally, (\ref{eq:layer-kfac}) follows readily from (\ref{eq:kfac-formula2}) by noticing that
$\calL_{\theta^n\theta^n} = \bar{\mA}_n \otimes \bar{\mB}_n$ under our computation.

\begin{remark}[Uncorrelated assumption of $\rvz^n \otimes \rvg^n_i$)]\normalfont
    This assumption is indeed strong yet almost necessary to yield tractable Kronecker matrices for efficient second-order operation. Tracing back to the development of Kronecker-based methods, similar assumptions also appear in convolution layers (\eg uncorrelated between spatial-wise derivatives \citep{grosse2016kronecker}) and recurrent units (\eg uncorrelated between temporal-wise derivatives \citep{martens2018kronecker}).
    The latter may be thought of as the discretization of Neural ODEs.
    We note, however, that it is possible to relax this assumption by considering tractable graphical models (\eg linear Gaussian \citep{martens2018kronecker}) at the cost of 2-3 times more operations per iteration. In terms of the performance difference, perhaps surprisingly, adopting tractable temporal models provides only minor improvement in test-time performance (see Fig.~4 in \citet{martens2018kronecker}). In some cases, it has been empirically observed that methods adopting the uncorrelated assumption yields better performance \citep{laurent2018evaluation}.
\end{remark}

\begin{remark}[Relation to Fisher Information Matrix]\normalfont
    Recall that for all experiments we apply Gaussian-Newton approximation to the terminal Hessian $\Qxx(t_1)$.
    This specific choice is partially based on empirical performance and computational purpose, yet it turns out that the resulting precondition matrices (\ref{eq:Quu}, \ref{eq:q-kfac}) can be interpreted as Fisher information matrix (FIM).
    In other words, under this specific setup, (\ref{eq:Quu}, \ref{eq:q-kfac}) can be equivalently viewed as the FIM of Neural ODEs.
    This implies SNOpt may be thought of as following Natural Gradient Descent (NGD), which is well-known for taking the steepest descent direction in the space of model distributions \citep{amari2000methods,martens2014new}.
    Indeed, it has been observed that NGD-based methods converge to equally good accuracies, even though its learning rate varies across 1-2 orders (see Fig 10 in \citet{ma2019inefficiency} and Fig 4 in \citet{george2018fast}). These observations coincide with our results (Fig.~\ref{fig:point-cloud}) for Neural ODEs.
\end{remark}

%% file: subtex/09appendix3.tex
\def\hT{{ \bar{T} }}
\def\hF{{ \bar{F} }}

\def\hPhi{{ \widetilde{\Phi} }}
\def\hPhixx{{ \hPhi_{\hvx\hvx} }}
\def\hPhixT{{ \hPhi_{\hvx\hT} }}
\def\hPhiTx{{ \hPhi_{\hT\hvx} }}
\def\hPhiTT{{ \hPhi_{\hT\hT} }}
\def\hPhiTxT{{ \hPhi_{\hT\hvx}^\T }}
\def\tQ{{\widetilde{Q}}}
\def\tQx{{\tQ_{\hvx}}}
\def\tQu{{\tQ_{\hvu}}}
\def\tQT{{\tQ_{\hT}}}

\def\tQxu{{\tQ_{\hvx\hvu}}}
\def\tQuu{{\tQ_{\hvu\hvu}}}
\def\tQux{{\tQ_{\hvu\hvx}}}
\def\tQuT{{\tQ_{\hvu\hT}}}
\def\tQTu{{\tQ_{\hT\hvu}}}

\def\tQxT{{\tQ_{\hvx\hT}}}
\def\tQTx{{\tQ_{\hT\hvx}}}
\def\tQxx{{\tQ_{\hvx\hvx}}}
\def\tQTT{{\tQ_{\hT\hT}}}

\textbf{Derivation of (\ref{eq:feedback}).}
Here we present an extension of our OCP framework to jointly optimizing the architecture of Neural ODEs,
specifically the integration bound $t_1$.
The proceeding derivation, despite being rather tedious, follows a similar procedure in Section~\ref{sec:3.1} and the proof of Theorem~\ref{prop:1}.

Recall the modified cost-to-go function that we consider for free-horizon optimization,
\begin{align*}
    \tQ(t,\rvx_t,\rvu_t, \mathrm{T}) := {\hPhi(\mathrm{T}, \rvx(\mathrm{T}))+ \int_t^{\mathrm{T}} \ell(\tau,\rvx_\tau,\rvu_\tau) \text{ } \rd \tau },
\end{align*}
where we introduce a new variable, \ie the terminal horizon $\mathrm{T}$, that shall be jointly optimized.
We use the expression $\rvx(\mathrm{T})$ to highlight the fact that the terminal state is now a function of $\mathrm{T}$.

Similar to what we have explored in Section~\ref{sec:3.1},
our goal is to derive an analytic expression for the derivatives of $\tQ$ at the integration start time $t_0$ w.r.t. this new variable $\mathrm{T}$.
This can be achieved by characterizing the local behavior of the following ODE,
\begin{align}
{0 = {\ell(t,\rvx_t,\rvu_t) + \fracdiff{\tQ(t,\rvx_t,\rvu_t, \mathrm{T})}{t}}}, \quad
\tQ({\mathrm{T}},\rvx_\mathrm{T}) = \hPhi(\mathrm{T}, \rvx(\mathrm{T})),
\label{eq:Q-ode-T}
\end{align}
expanded on some nominal solution path $(\hvx_t,\hvu_t,\hT)$.

Let us start from the terminal condition in (\ref{eq:Q-ode-T}).
Given $\tQ(\hT, \hvx_{\hT}) = \hPhi(\hT, \hvx(\hT))$,
perturbing the terminal horizon $\hT$ by an infinitesimal amount $\delta \mathrm{T}$ yields
\begin{align}
\begin{split}
    \tQ(\hT + \delta \mathrm{T}, \hvx_{\hT + \delta \mathrm{T}})
    &= \ell(\hvx_\hT,\hvu_\hT) \delta \mathrm{T} +  \hPhi(\hT + \delta \mathrm{T}, \hvx(\hT + \delta \mathrm{T})).
    \label{eq:dQ-T-app}
\end{split}
\end{align}
It can be shown that the second-order expansion of the last term in (\ref{eq:dQ-T-app})
takes the form,
\begin{align}
\begin{split}
    \hPhi\left(\hT + \delta \mathrm{T}, \hvx(\hT + \delta \mathrm{T})\right)
    &\approx
    \hPhi\left(\hT, \hvx(\hT)\right)
    + \hPhi_\hvx^\T \delta \rvx_\hT
    + \left( \hPhi_\hT  + \hPhi_\hvx^\T \hF \right) \delta \mathrm{T}
    + \frac{1}{2} \delta \rvx_\hT^\T \hPhixx\delta \rvx_\hT \\
    &\quad +
    \frac{1}{2} \delta \rvx_\hT^\T \left( \hPhixT + \hPhixx\hF \right) \delta \mathrm{T} +
    \frac{1}{2} \delta \mathrm{T} \left( \hPhiTx + \hF^\T\hPhixx \right) \delta \rvx_\hT
    \\
    &\quad +
    \frac{1}{2} \delta \mathrm{T}  \left(
        \hPhiTT + \hPhiTx\hF + \hF^\T\hPhixT + \hF^\T\hPhixx \hF
    \right) \delta \mathrm{T},
\label{eq:phi-T-expand}
\end{split}
\end{align}
which relies on the fact that the following formula holds for any generic function that takes $t$ and $\rvx(t)$ as its arguments:
\begin{align*}
    \fracdiff{}{t}(\cdot)
    = \fracpartial{}{t}(\cdot) + \fracpartial{}{\rvx}(\cdot)^\T \hF,
    \quad \text{where } \hF = F(t, \hvx_t, \hvu_t).
\end{align*}

Substituting (\ref{eq:phi-T-expand}) to (\ref{eq:dQ-T-app}) gives us the
local expressions of the terminal condition up to second-order,
\begin{subequations}
\begin{alignat}{2}
\tQx(\hT) &= \hPhi_\hvx, \quad
&&\text{ }\text{ } \tQT(\hT) = \ell(\hvx_\hT,\hvu_\hT) + \hPhi_\hT + \hPhi_\hvx^\T \hF ,  \\
\tQTx(\hT) &= \hPhiTx + \hF^\T\hPhixx, \quad
&& \tQxT(\hT) = \hPhixT + \hPhixx\hF, \\
\tQxx(\hT) &= \hPhi_{\hvx\hvx}, \quad
&& \tQTT(\hT) = \hPhiTT + \hPhiTxT\hF + \hF^\T\hPhixT + \hF^\T\hPhixx \hF,
\end{alignat} \label{eq:cddp-qT-term}
\end{subequations}
where $ \tQx(\hT) \equiv \frac{\delta \tQ }{\delta \rvx_\hT} = \frac{\tQ(\hT + \delta \mathrm{T}, \hvx_{\hT + \delta \mathrm{T}}) - \tQ(\hT, \hvx_{\hT}) }{\delta \rvx_\hT} $, and etc.

Next, consider the ODE dynamics in (\ref{eq:Q-ode-T}).
Similar to (\ref{eq:q-expansion}), we can
expand $\tQ$ w.r.t. all optimizing variables, \ie ($\rvx_t$, $\rvu_t$, $\mathrm{T}$),
up to second-order.
In this case, the approximation is given by
\begin{equation}
\begin{split}
\tQ(t, \hvx_t,\hvu_t,\hT) +
\tQx^\T \drvxt + \tQu^\T \drvut + \markgreen{\tQT \delta \mathrm{T}} +
\frac{1}{2} {\bvec{\drvxt \\[0.5ex] \drvut \\[0.5ex] \markgreen{\delta \mathrm{T}} }}^\T
\begin{bmatrix}
    \tQxx & \tQxu & \markgreen{\tQxT} \\[0.5ex]
    \tQux & \tQuu & \markgreen{\tQuT} \\[0.5ex]
    \markgreen{\tQTx} & \markgreen{\tQTu} & \markgreen{\tQTT}
\end{bmatrix}
\bvec{\drvxt \\[0.5ex] \drvut \\[0.5ex] \markgreen{\delta \mathrm{T}}},
\end{split} \label{eq:QT-expand}
\end{equation}
which shares the same form as (\ref{eq:q-expansion}) except having additional terms that account for the derivatives related to $\mathrm{T}$ (\markgreen{marked as green}).
Substitute (\ref{eq:QT-expand}) to the ODE dynamics in (\ref{eq:Q-ode-T}),
then expand the time derivatives $\fracdiff{}{t}$ as in (\ref{eq:Q-2nd-expand}),
and finally replace $\fracdiff{\drvxt}{t}$, $\fracdiff{\drvut}{t}$, and $\fracdiff{\delta \mathrm{T}}{t}$ with
\begin{align*}
\fracdiff{\drvxt}{t} = {\FxT \drvxt + \FuT \drvut}, \quad
\fracdiff{\drvut}{t} = \mathbf{0}, \quad \text{ and }
 \fracdiff{\delta \mathrm{T}}{t} = 0.
\end{align*}
Then, it can be shown that the first and second-order derivatives of $\tQ$ w.r.t. $\mathrm{T}$ obey the following backward ODEs:
\begin{align*}
- \fracdiff{\tQT}{t} = 0, \quad
- \fracdiff{\tQTT}{t} = 0, \quad
- \fracdiff{\tQTx}{t} = \tQTx \Fx, \quad
- \fracdiff{\tQTu}{t} = \tQTx \Fu,
\end{align*}
with the terminal condition given by (\ref{eq:cddp-qT-term}).
As for the derivatives that do not involve $\mathrm{T}$, \eg  $\tQxx$ and $\tQuu$, one can verify that they follow the same backward structures
given in (\ref{eq:cddp-q}) except changing the terminal condition from $\Phi$ to $\hPhi$.

To summarize, solving the following ODEs gives us the derivatives of $\tQ$ related to $\mathrm{T}$ at $t_0$:
\begin{subequations}
\begin{empheq}[box=\widefbox]{align}
  - \fracdiff{}{t} \tQT(t)  &= 0, \qquad\qquad\text{ }\text{ } \tQT(\hT) = \ell(\hvx_\hT,\hvu_\hT) + \hPhi_\hT + \hPhi_\hvx^\T \hF \label{eq:T-ode-a} \\
  - \fracdiff{}{t} \tQTT(t) &= 0, \qquad\quad\text{ }\text{ }\text{ } \tQTT(\hT) = \hPhiTT + \hPhiTxT\hF + \hF^\T\hPhixT + \hF^\T\hPhixx \hF \label{eq:T-ode-b} \\
  - \fracdiff{}{t} \tQTx(t) &= \tQTx \Fx, \quad\text{ } \tQTx(\hT) = \hPhiTx + \hF^\T\hPhixx \label{eq:T-ode-c} \\
  - \fracdiff{}{t} \tQTu(t) &= \tQTx \Fu, \quad\text{ } \tQTu(\hT) = \mathbf{0} \label{eq:T-ode-d}
\end{empheq} \label{eq:T-ode}
\end{subequations}
Then,
we can consider the following quadratic programming for the optimal perturbation $\delta \mathrm{T}^*$,
\begin{align*}
\min_{\delta \mathrm{T}} \text{ }&
\Qx(t_0)^\T \drvx_{t_0} + \Qu(t_0)^\T \drvu_{t_0} +
{\tQT(t_0) \delta \mathrm{T}} \\ &\quad+
\frac{1}{2} {\bvec{\drvx_{t_0} \\[0.3ex] \drvu_{t_0} \\[0.3ex] {\delta \mathrm{T}} }}^\T
\begin{bmatrix}
    \tQxx(t_0) & \tQxu(t_0) & {\tQxT(t_0)} \\[0.3ex]
    \tQux(t_0) & \tQuu(t_0) & {\tQuT(t_0)} \\[0.3ex]
    {\tQTx(t_0)} & {\tQTu(t_0)} & {\tQTT(t_0)}
\end{bmatrix}
\bvec{\drvx_{t_0} \\[0.3ex] \drvu_{t_0} \\[0.3ex] {\delta \mathrm{T}}},
\end{align*}
which has an analytic feedback solution given by
\begin{empheq}[box=\widefbox]{align*}
    \delta \mathrm{T}^*(\drvx_{t_0}, \drvu_{t_0})
     = [\tQTT(t_0)]^{-1}\left( \tQT(t_0) + \tQTx(t_0)\drvx_{t_0} + \tQTu(t_0)\drvu_{t_0} \right).
\end{empheq}
In practice, we drop the state differential $\drvx_{t_0}$ and only keep the control differential $\drvu_{t_0}$,
which can be viewed as the parameter update $\delta \theta$ by recalling (\ref{eq:ct-ocp}).
With these, we arrive at
the second-order feedback policy presented in (\ref{eq:feedback}).

\textbf{Practical implementation.}
We consider a vanilla quadratic cost,
$\widetilde{\Phi}(\mathrm{T}, \rvx(\mathrm{T})) := {\Phi}(\rvx(\mathrm{T})) + \frac{c}{2}\mathrm{T}^2$,
which penalizes longer integration time,
and impose Gaussian-Newton approximation for the terminal cost, \ie ${\Phi}_{\hvx\hvx} \approx \Phi_{\hvx}\Phi_{\hvx}^\T $.
With these, the terminal conditions in (\ref{eq:T-ode}) can be simplified to
\begin{align*}
\tQT(\hT) = c\hT + \Phi_{\hvx}^\T \hF, \quad
\tQTT(\hT) = c + \left(\Phi_{\hvx}^\T \hF\right)^2, \quad
\tQTx(\hT) = \left(\Phi_{\hvx}^\T \hF\right) \Phi_{\hvx}^\T.
\end{align*}
Since $\tQT(t)$ and $\tQTT(t)$ are time-invariant (see (\ref{eq:T-ode-a}, \ref{eq:T-ode-b})),
we know the values of $\tQT(t_0)$ and $\tQTT(t_0)$ at the terminal stage.
Further, one can verify that $\forall t\in [t_0, \hT], \quad \tQTu(t) = \left(\Phi_{\hvx}^\T \hF\right) \Qu(t)^\T$.
In other words, the feedback term $\tQTu$ simply rescales the first-order derivative $\tQu$ by $\Phi_{\hvx}^\T \hF$.
These reasonings suggest that we can evaluate the second-order feedback policy (\ref{eq:feedback}) almost at no cost
without augmenting any additional state to \texttt{\markblue{ODESolve}}.
Finally, to adopt the stochastic training, we keep the moving averages of all terms and update $\mathrm{T}$ with (\ref{eq:feedback}) every 50-100 training iterations.

%% file: subtex/10appendix4.tex
All experiments are conducted on the same GPU machine (TITAN RTX)
and implemented with \texttt{pytorch}.
Below we provide full discussions on topics that are deferred from Section~\ref{sec:4}.

\textbf{Model configuration.}
  Here, we specify the model for each dataset.
  We will adopt the following syntax to describe the layer configuration.
  \begin{itemize}[leftmargin=20pt]
  \item
  {\small \texttt{Linear(input\_dim, output\_dim)}}
  \item {\small \texttt{Conv(output\_channel, kernel, stride)}}
  \item {\small \texttt{ConcatSquashLinear(input\_dim, output\_dim)}}\footnote{
      {\fontsize{8pt}{8pt}\selectfont \url{https://github.com/rtqichen/ffjord/blob/master/lib/layers/diffeq_layers/basic.py\#L76}}
  }
  \item {\small \texttt{GRUCell(input\_dim, hidden\_dim)}}
  \end{itemize}
  Table~\ref{table:F-config} details the vector field $F(t,\rvx_t,\theta)$ of Neural ODEs for each dataset.
  All vector fields are represented by some DNNs,
  and their architectures are adopted from previous references as listed.
  The convolution-based feature extraction of image-classification models consists of 3 convolution layers connected through $\texttt{ReLU}$, \ie
  { \small $\texttt{Conv(64,3,1)} \rightarrow \texttt{ReLU} \rightarrow \texttt{Conv(64,4,2)} \rightarrow \texttt{ReLU} \rightarrow \texttt{Conv(64,4,2)}$}.
  For time-series models,
  We set the dimension of the hidden space to 32, 64, and 32 respectively for SpoAD, ArtWR, and CharT.
  Hence, their GRU cells are configured by {\small \texttt{GRUCell(27,32)}}, {\small \texttt{GRUCell(19,64)}}, and {\small \texttt{GRUCell(7,32)}}.
  Since these models take regular time-series with the interval of 1 second,
  the integration intervals of their Neural ODEs are set to $\{0,1,\cdots,K\}$, where $K$ is the series length listed in Table~\ref{table:1}.
  Finally,
  we find that using 1 Neural ODE is sufficient to achieve good performance on Circle and Miniboone,
  whereas for Gas, we use 5 Neural ODEs stacked in sequence.

\input{subtex/12figure}

\textbf{Tuning process.}
  We perform a grid search on tuning the hyper-parameters (\eg learning rate, weight decay) for each method on each dataset.
  The search grid for each method is detailed in Table~\ref{table:hyper}.
  All figures and tables mentioned in Section~\ref{sec:4} report the best-tuned results.
  For time-series models,
  we employ standard learning rate decay
  and note that without this annealing mechanism, we are unable to have first-order baselines converge stably.
  We also observe that the magnitude of the gradients of the GRU cells is typically 10-50 larger than the one of the Neural ODEs.
  This can make training unstable when the same configured optimizer is used to train all modules.
  Hence,
  in practice
  we fix Adam to train the GRUs while varying the optimizer for training Neural ODEs.
  Lastly, for image classification models, we deploy our method together with
  the standard Kronecker-based method \citep{grosse2016kronecker} for training the convolution layers.
  This enables full second-order training for the entire model,
  where the Neural ODE, as a continuous-time layer, is trained using our method proposed in Alg.~\ref{alg:1}.
  Finally, the momentum value for SGD is set to 0.9.

\textbf{Dataset.}
  All image datasets are preprocessed with standardization.
  To accelerate training, we utilize 10\% of the samples in Gas,
  which still contains  85,217 training samples and 10,520 test samples.
  In general, the relative performance among training
  methods remains consistent
  for larger dataset ratios.

\textbf{Setup and motivation of Fig.~\ref{fig:adap-t1}.}
  We initialize all Neural ODEs with the \textit{same} parameters
  while only varying the integration bound $t_1$.
  By manually grid-searching over $t_1$,
  Fig.~\ref{fig:adap-t1} implies that despite initializing from the same parameter, different $t_1$
  can yield distinct training time and accuracy;
  in other words, different $t_1$ can lead to distinct ODE solution.
  As an ideal Neural ODE model should keep the training time as small as possible without sacrificing the accuracy, there is a clear motivation to adaptive/optimize $t_1$ throughout training.
  Additional comparison w.r.t. standard (\ie static) residual models can be founded in Appendix~\ref{app:5}.

\textbf{Generating Fig.~\ref{fig:complexity}.}
  The numerical values of the per-iteration runtime are reported in Table~\ref{table:runtime},
  whereas the ones for the memory consumption are given in Table~\ref{table:memory}.
  We use the last rows (\ie $\frac{\text{SNOpt}}{\text{Adam}}$) of these two tables to generate Fig.~\ref{fig:complexity}.

\input{subtex/13figure}

\begin{minipage}{0.42\textwidth}
\textbf{Tikhonov regularization in line 10 of Alg.~\ref{alg:1}.}
  In practice,
  we apply Tikhonov regularization to the precondition matrix, \ie $\calL_{\theta^n\theta^n} + \epsilon\mI$,
  where $\theta^n$ is the parameter of layer $n$ (see Fig.~\ref{fig:3} and (\ref{eq:q-kfac}))
  and $\epsilon$ is the Tikhonov regularization widely used for stabilizing second-order training \citep{botev2017practical,zhang2019fast}.
  To efficiently compute this $\epsilon$-regularized Kronecker precondition matrix without additional factorization
  or approximation (\eg Section 6.3 in \citet{martens2015optimizing}),
  we instead follow \citet{george2018fast} and
  \end{minipage}
  \hfill
  \begin{minipage}{0.55\textwidth}
  \vskip -0.15in
  \begin{algorithm}[H]
   \caption{$\epsilon$-regularized Kronecker Update} %
   \label{alg:2}
   \begin{algorithmic}[1]
   \STATE {\bfseries Input:}
          Tikhonov regularization $\epsilon$,
          amortization $\alpha$, \\[0.3ex]
          $\quad\qquad$Kronecker matrices $\bar{\mA}_n$ $\bar{\mB}_n$

   \STATE $\mU_\mA, \Sigma_\mA =$ \texttt{EigenDecomposition(}$\bar{\mA}_n$\texttt{)}
   \STATE $\mU_\mB, \Sigma_\mB =$ \texttt{EigenDecomposition(}$\bar{\mB}_n$\texttt{)}
   \STATE $\mX :=$
          \texttt{vec}$^{-1}$\texttt{(}$(\mU_\mA \otimes \mU_\mB)^\T \calL_{\theta^n}$\texttt{)}
          $=\mU_\mB^\T \widetilde{\calL}_{\theta^n} \mU_\mA$
   \STATE $\mS^* := \alpha \mS^* + (1-\alpha) \mX^2$
   \STATE $\mX := \mX / (\mS^* + \epsilon)$
   \STATE $\delta\theta := (\mU_\mA \otimes \mU_\mB)$\texttt{vec(}$\mX$\texttt{)}
          $=$\texttt{vec(}$\mU_\mB\mX\mU_\mA^\T$\texttt{)}
   \STATE $\theta \leftarrow \theta - \eta \delta\theta$
   \end{algorithmic}
  \end{algorithm}
  \end{minipage}
  perform eigen-decompositions,
  \ie $\bar{\mA}_n = \mU_\mA \Sigma_\mA \mU_\mA^\T$ and $\bar{\mB}_n = \mU_\mB \Sigma_\mB \mU_\mB^\T$,
  so that we can utilize the property of Kronecker product \citep{schacke2004kronecker} to obtain
  \begin{align}
    (\bar{\mA}_n + \bar{\mB}_n + \epsilon\mI)^{-1}
    = (\mU_\mA \otimes \mU_\mB) (\Sigma_\mA \otimes \Sigma_\mB + \epsilon)^{-1} (\mU_\mA \otimes \mU_\mB)^\T.
    \label{eq:tik-reg}
  \end{align}
  This, together with the eigen-based amortization which substitutes the original diagonal matrix
  $\mS := \Sigma_\mA \otimes \Sigma_\mB$
  in (\ref{eq:tik-reg}) with $\mS^* := ((\mU_\mA \otimes \mU_\mB)^\T \calL_{\theta^n})^2$,
  leads to the computation in Alg.~\ref{alg:2}.
  Note that \texttt{vec} is the shorthand for vectorization, and we denote $\calL_{\theta^n} = $\texttt{vec(}$\widetilde{\calL}_{\theta^n}$\texttt{)}. Finally, $\alpha$ is the amortizing coefficient, which we set to 0.75 for all experiments. As for $\epsilon$, we test 3 different values from \{0.1, 0.05, 0.03\} and report the best result.

\textbf{Error bar in Table~\ref{table:accu}.}
  Table~\ref{table:accu-bar} reports the standard derivations of Table~\ref{table:accu},
  indicating that our result remains statistically sound with comparatively lower variance.

  \newcommand{\bbar}[2]{{{\fontsize{7}{8}\selectfont#1$\pm$#2}}}

  \begin{figure}[H]
    \vskip -0.05in
    \centering
    \setlength\tabcolsep{3pt}
    \captionsetup{type=table}
    \caption{Test-time performance: accuracies for {\color{label1}image} and {\color{label3}time-series} datasets; NLL for {\color{label2}CNF} datasets}
    \vskip -0.03in
    \centering
    \begin{tabular}{lccccccccc}
      \toprule
      & {MNIST}  & {SVHN} & {CIFAR10}
      & {SpoAD}  & {ArtWR} & {CharT}
      & Circle & {Gas}  & {Minib.} \\
      \addlinespace[-0.25em]\arrayrulecolor{label1}
      \cmidrule[1pt](lr){2-4}\corcmidrule\arrayrulecolor{label3}%
      \cmidrule[1pt](lr){5-7}\corcmidrule\arrayrulecolor{label2}%
      \cmidrule[1pt](lr){8-10}\corcmidrule
      \addlinespace[0.4em]\arrayrulecolor{black}
      \midrule
      Adam  &  \bbar{98.83}{0.18}  &  \bbar{91.92}{0.33}  &  \bbar{77.41}{0.51}
            &  \bbar{94.64}{1.12}  &  \bbar{84.14}{2.53}  & \bbar{93.29}{1.59}
            &  \bbar{0.90}{\textbf{0.02}}   &  \bbar{-6.42}{\textbf{0.18}}  & \bbar{13.10}{0.33} \\[2pt]

      SGD   &  \bbar{98.68}{0.22}          & \bbar{93.34}{1.17}          & \bbar{76.42}{0.51}
            &  \bbar{\textbf{97.70}}{0.69} & \bbar{85.82}{3.83}          & \bbar{95.93}{0.22}
            &  \bbar{0.94}{0.03}           & \bbar{-4.58}{0.23}          &  \bbar{13.75}{0.19} \\[1pt]
      \midrule

      \textbf{SNOpt}
            &  \bbar{\textbf{98.99}}{\textbf{0.15}} &   \bbar{\textbf{95.77}}{\textbf{0.18}} &  \bbar{\textbf{79.11}}{\textbf{0.48}}
            &  \bbar{97.41}{\textbf{0.46}} &  \bbar{\textbf{90.23}}{\textbf{1.49}} &  \bbar{\textbf{96.63}}{\textbf{0.19}}
            &  \bbar{\textbf{0.86}}{0.04}  &  \bbar{\textbf{-7.55}}{0.46} &   \bbar{\textbf{12.50}}{\textbf{0.12}} \\
      \bottomrule
    \end{tabular} \label{table:accu-bar}
  \end{figure}

\textbf{Discussion on Footnote \ref{footnote:4}.}
  Here, we provide some reasoning on why the preconditioned updates may lead the parameter to regions that are stabler for integration.
  We first adopt the theoretical results in \citet{martens2015optimizing}, particularly their Theorem 1 and Corollary 3, to our setup.
  \begin{corollary}[Preconditioned Neural ODEs] \label{coro:4}
          Updating the parameter of a Neural ODE, $F(\cdot, \cdot, \theta)$, with the preconditioned updates in (\ref{eq:layer-kfac})
          is equivalent to updating the parameter $\theta^\dagger \in \mathbb{R}^n$ of a ``preconditioned'' Neural ODE, $F^\dagger(\cdot, \cdot, \theta^\dagger)$, with gradient descent.
          This preconditioned Neural ODE has all the activations $\rvz^n$ and derivatives $F_{\rvh^n}^\T \rvq_i$ (see Fig.~\ref{fig:3}) centered and whitened.
  \end{corollary}
  These centering and whitening mechanisms are known to enhance convergence \citep{desjardins2015natural} and closely relate to Batch Normalization \citep{ioffe2015batch},
  which effectively smoothens the optimization landscape \citep{santurkar2018does}.
  Hence, one shall expect it also smoothens the diffeomorphism of both the forward and backward ODEs (\ref{eq:node}, \ref{eq:backward-ode}) of Neural ODEs.

%% file: subtex/12figure.tex
\begin{figure}[t]
  \vskip -0.1in
  \begin{minipage}{\textwidth}
  \centering
  \captionsetup{type=table}
  \captionsetup{justification=centering}
  \caption{
      Configuration of the vector field $F(t,\rvx_t, \theta)$ of Neural ODEs used for each dataset \\
      ($^\ddagger$MIT License; $^\S$Apache License)}
  \vskip -0.05in
  \centering
  \begin{tabular}{lll}
  \toprule
  Dataset & DNN architecture as $F(t,\rvx_t, \theta)$ & Model reference   \\
  \midrule
  \specialcelll{MNIST \\ SVHN \\ CIFAR10}
    &
        { \small $\texttt{Conv(64,3,1)} \rightarrow \texttt{ReLU} \rightarrow \texttt{Conv(64,3,1)}$}
    &   \citet{chen2018neural}$^\ddagger$
    \\[1ex]
  \midrule
  \specialcelll{SpoAD \\ CharT}
    &   { \small $\begin{aligned}
                &\texttt{Linear(32,32)} \rightarrow \texttt{Tanh} \rightarrow \texttt{Linear(32,32)} \\
                &\rightarrow \texttt{Tanh} \rightarrow \texttt{Linear(32,32)} \rightarrow \texttt{Tanh} \\
                &\rightarrow \texttt{Linear(32,32)}
        \end{aligned}$}
    &   \citet{kidger2020neural}$^\S$
    \\[1ex]
  \midrule
  ArtWR
    &   { \small $\begin{aligned}
                &\texttt{Linear(64,64)} \rightarrow \texttt{Tanh} \rightarrow \texttt{Linear(64,64)} \\
                &\rightarrow \texttt{Tanh} \rightarrow \texttt{Linear(64,64)} \rightarrow \texttt{Tanh} \\
                &\rightarrow \texttt{Linear(64,64)}
        \end{aligned}$}
    &   \citet{kidger2020neural}$^\S$
    \\[1ex]
  \midrule
  Circle
    &   { \small $\texttt{Linear(2,64)}^{\text{\scriptsize \ref{foot:5}}} \rightarrow \texttt{Tanh} \rightarrow \texttt{Linear(64,2)}$}\tablefootnote{
            The weights of both \texttt{Linear} layers are generated from a \texttt{HyperNet} implemented in
            {\fontsize{8pt}{8pt}\selectfont \url{https://github.com/rtqichen/torchdiffeq/blob/master/examples/cnf.py\#L77-L114}}. \label{foot:5}
        }
    &   \citet{chen2018neural}$^\ddagger$
    \\[1ex]
  \midrule
  Gas
    &   { \small $\begin{aligned}
                &\texttt{ConcatSquashLinear(8,160)} \rightarrow \texttt{Tanh} \\
                &\rightarrow \texttt{ConcatSquashLinear(160,160)} \rightarrow \texttt{Tanh} \\
                &\rightarrow \texttt{ConcatSquashLinear(160,160)} \rightarrow \texttt{Tanh} \\
                &\rightarrow \texttt{ConcatSquashLinear(160,8)}
        \end{aligned}$}
    &   \citet{grathwohl2018ffjord}$^\ddagger$
    \\[1ex]
  \midrule
  Miniboone
    &   { \small $\begin{aligned}
                &\texttt{ConcatSquashLinear(43,860)} \rightarrow \texttt{SoftPlus} \\
                &\rightarrow \texttt{ConcatSquashLinear(860,860)} \\
                &\rightarrow \texttt{SoftPlus} \rightarrow \texttt{ConcatSquashLinear(860,43)}
        \end{aligned}$}
    &   \citet{grathwohl2018ffjord}$^\ddagger$
    \\[1ex]
  \bottomrule
  \end{tabular} \label{table:F-config}
  \end{minipage}
  \vskip 0.15in
  \begin{minipage}{\textwidth}
    \setlength\tabcolsep{3pt}
    \centering
    \captionsetup{type=table}
    \caption{
        Hyper-parameter grid search considered for each method
    }
    \vskip -0.05in
    \centering
    \begin{tabular}{lcc}
      \toprule
      Method & Learning rate & Weight decay \\
      \midrule
      Adam  & \{ 1e-4, 3e-4, 5e-4, 7e-4, 1e-3, 3e-3, 5e-3, 7e-3, 1e-2, 3e-2, 5e-2 \} & \{0.0, 1e-4, 1e-3 \} \\
      SGD   & \{ 1e-3, 3e-3, 5e-3, 7e-3, 1e-2, 3e-2, 5e-2, 7e-2, 1e-1, 3e-1, 5e-1 \} & \{0.0, 1e-4, 1e-3 \} \\
      \textbf{Ours} & \{ 1e-3, 3e-3, 5e-3, 7e-3, 1e-2, 3e-2, 5e-2, 7e-2, 1e-1, 3e-1, 5e-1 \} & \{0.0, 1e-4, 1e-3 \} \\
      \bottomrule
    \end{tabular} \label{table:hyper}
    \vskip -0.1in
  \end{minipage}
  \vskip -0.1in
\end{figure}

%% file: subtex/13figure.tex
\begin{figure}[H]
  \vskip -0.05in
  \begin{minipage}{\textwidth}
    \centering
    \setlength\tabcolsep{5pt}
    \captionsetup{type=table}
    \caption{Per-iteration runtime (seconds) of different optimizers on each dataset}
    \vskip -0.05in
    \centering
    \begin{tabular}{lccccccccc}
      \toprule
      &
      \multicolumn{3}{c}{Image Classification} &
      \multicolumn{3}{c}{Time-series Prediction} &
      \multicolumn{3}{c}{Continuous NF}        \\
      \arrayrulecolor{label1}
      \cmidrule[1pt](lr){2-4}\corcmidrule\arrayrulecolor{label3}%
      \cmidrule[1pt](lr){5-7}\corcmidrule\arrayrulecolor{label2}%
      \cmidrule[1pt](lr){8-10}\corcmidrule
      \addlinespace[0.6em]
      \arrayrulecolor{black}
      & MNIST     & SVHN     & CIFAR10 & SpoAD & ArtWR & CharT & Circle & Gas & Minib. \\
      \midrule
      Adam          & 0.15 &  0.78 &  0.17 &  5.24 &  9.95  & 14.79 &  0.34 &  2.25 &  0.65 \\
      SGD           & 0.15 &  0.81 &  0.17 &  5.23 &  10.00 & 14.77 &  0.33 &  2.28 &  0.74 \\
      \textbf{SNOpt} & 0.15 &  0.68 &  0.20 &  5.18 &  10.05 & 14.89 &  0.94 &  4.34 &  1.04 \\
      \midrule
      $\frac{\text{SNOpt}}{\text{Adam}}$
                    & 1.00 &  0.87 &  1.16 &  0.99 &  1.01 & 1.01 &  2.75 &  1.93 &  1.60 \\
      \bottomrule
    \end{tabular} \label{table:runtime}
  \end{minipage}
  \vskip 0.2in
  \begin{minipage}{\textwidth}
    \centering
    \setlength\tabcolsep{5pt}
    \captionsetup{type=table}
    \caption{Memory Consumption (GBs) of different optimizers on each dataset}
    \vskip -0.05in
    \centering
    \begin{tabular}{lccccccccc}
      \toprule
      &
      \multicolumn{3}{c}{Image Classification} &
      \multicolumn{3}{c}{Time-series Prediction} &
      \multicolumn{3}{c}{Continuous NF}        \\
      \arrayrulecolor{label1}
      \cmidrule[1pt](lr){2-4}\corcmidrule\arrayrulecolor{label3}%
      \cmidrule[1pt](lr){5-7}\corcmidrule\arrayrulecolor{label2}%
      \cmidrule[1pt](lr){8-10}\corcmidrule
      \addlinespace[0.6em]
      \arrayrulecolor{black}
      & MNIST     & SVHN     & CIFAR10 & SpoAD & ArtWR & CharT & Circle & Gas & Minib. \\
      \midrule
      Adam          & 1.23 &  1.29 &  1.29 &  1.39 &  1.18 & 1.24 &  1.13 &  1.17 &  1.28 \\
      SGD           & 1.23 &  1.28 &  1.28 &  1.39 &  1.18 & 1.24 &  1.13 &  1.17 &  1.28 \\
      \textbf{SNOpt} & 1.64 &  1.81 &  1.81 &  1.49 &  1.28 & 1.34 &  1.15 &  1.34 &  1.68 \\
      \midrule
      $\frac{\text{SNOpt}}{\text{Adam}}$
                    & 1.33 &  1.40 &  1.40 &  1.07 &  1.09 & 1.08 &  1.02 &  1.14 &  1.31 \\
      \bottomrule
    \end{tabular} \label{table:memory}
  \end{minipage}
\end{figure}

%% file: subtex/11appendix5.tex
\begin{figure}[H]
    \vskip -0.1in
    \begin{minipage}{\textwidth}
      \centering
      \includegraphics[height=2.7cm]{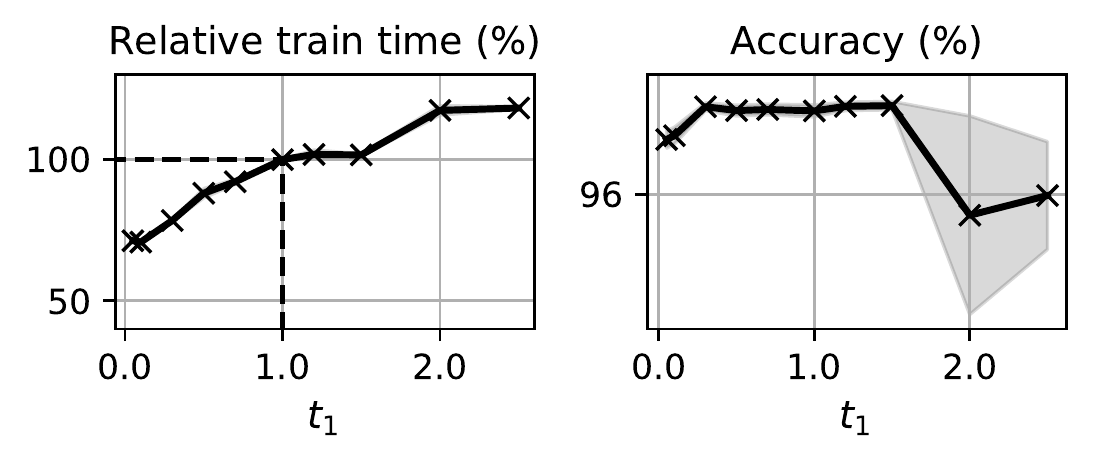}
      \vskip -0.05in
      \caption{
        Training performance of MNIST with Adam when using different $t_1$.
      }
      \label{fig:app-adap-t1a}
    \end{minipage}
    $\text{ }\text{ }$
    \begin{minipage}{\textwidth}
      \vskip 0.15in
      \centering
      \subfloat{
        \includegraphics[width=0.9\textwidth]{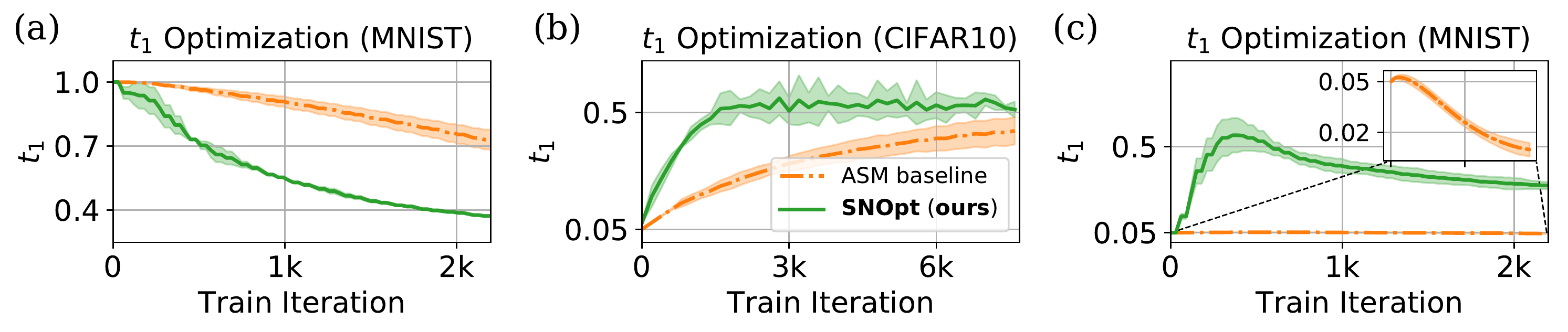}
        \label{fig:app-adap-t1b-a}
      }
      \subfloat{
        \textcolor{white}{\rule{1pt}{1pt}}
        \label{fig:app-adap-t1b-b}
      }
      \subfloat{
        \textcolor{white}{\rule{1pt}{1pt}}
        \label{fig:app-adap-t1b-c}
      }
      \vskip -0.05in
      \caption{
        Dynamics of $t_1$ over training using different methods,
        where we consider
        (a) MNIST training with $t_1$ initialized to 1.0, and
        (b, c) CIFAR10 and MNIST training with $t_1$ initialized to some unstable small values (\eg 0.05).
      }
      \label{fig:app-adap-t1b}
    \end{minipage}
    \vskip -0.1in
\end{figure}

\textbf{$t_1$ optimization.}
    Fig.~\ref{fig:app-adap-t1a} shows that a similar behavior (as in Fig.~\ref{fig:adap-t1}) can be found when training MNIST:
    while the accuracy remains almost stationary as we decrease $t_1$ from $1.0$,
    the required training time can drop by 20-35\%.
    Finally,
    we provide additional experiments for $t_1$ optimization in Fig.~\ref{fig:app-adap-t1b}.
    Specifically, Fig.~\ref{fig:app-adap-t1b-a} repeats the same experiment (as in Fig.~\ref{fig:adap-t1b}) on training MNIST,
    showing that our method (green curve) converges faster than the baseline.
    Meanwhile, Fig.~\ref{fig:app-adap-t1b-b} and \ref{fig:app-adap-t1b-c} suggest that our approach is also more effective in
    recovering from an unstable initialization of $t_1$.
    Note that both Fig.~\ref{fig:adap-t1b} and ~\ref{fig:app-adap-t1b} use Adam to optimize the parameter $\theta$.

\begin{figure}[H]
    \vskip -0.2in
    \begin{minipage}{\textwidth}
        \begin{center}
        \captionsetup{type=figure}
        \includegraphics[width=0.97\textwidth]{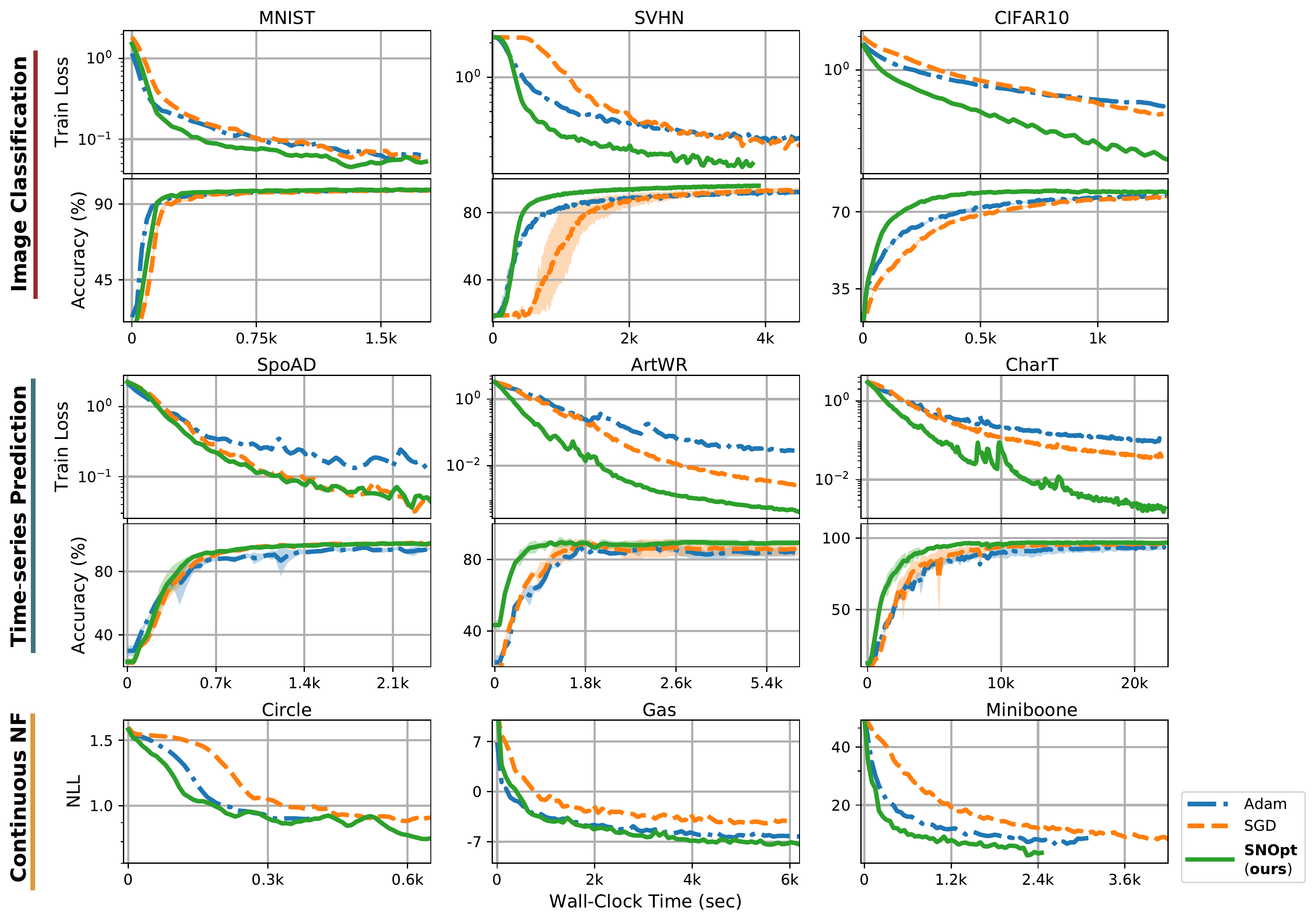}
        \vskip -0.05in
        \caption{
        Optimization performance measured by \emph{wall-clock} time
        across 9 datasets, including image (1$^\text{st}$-2$^\text{nd}$ rows) and time-series (3$^\text{rd}$-4$^\text{th}$ rows) classification, and continuous NF (5$^\text{th}$ row).
        We repeat the same figure with update iterations as x-axes in Fig~\ref{fig:convergence-full2}.
        Our method (green) achieves faster convergence rate compared to first-order baselines.
        Each curve is averaged over 3 random trials.
        }
        \label{fig:convergence-full}
        \end{center}
    \end{minipage}
    \vskip 0.1in
    \begin{minipage}{\textwidth}
        \begin{center}
        \captionsetup{type=figure}
        \includegraphics[width=0.97\textwidth]{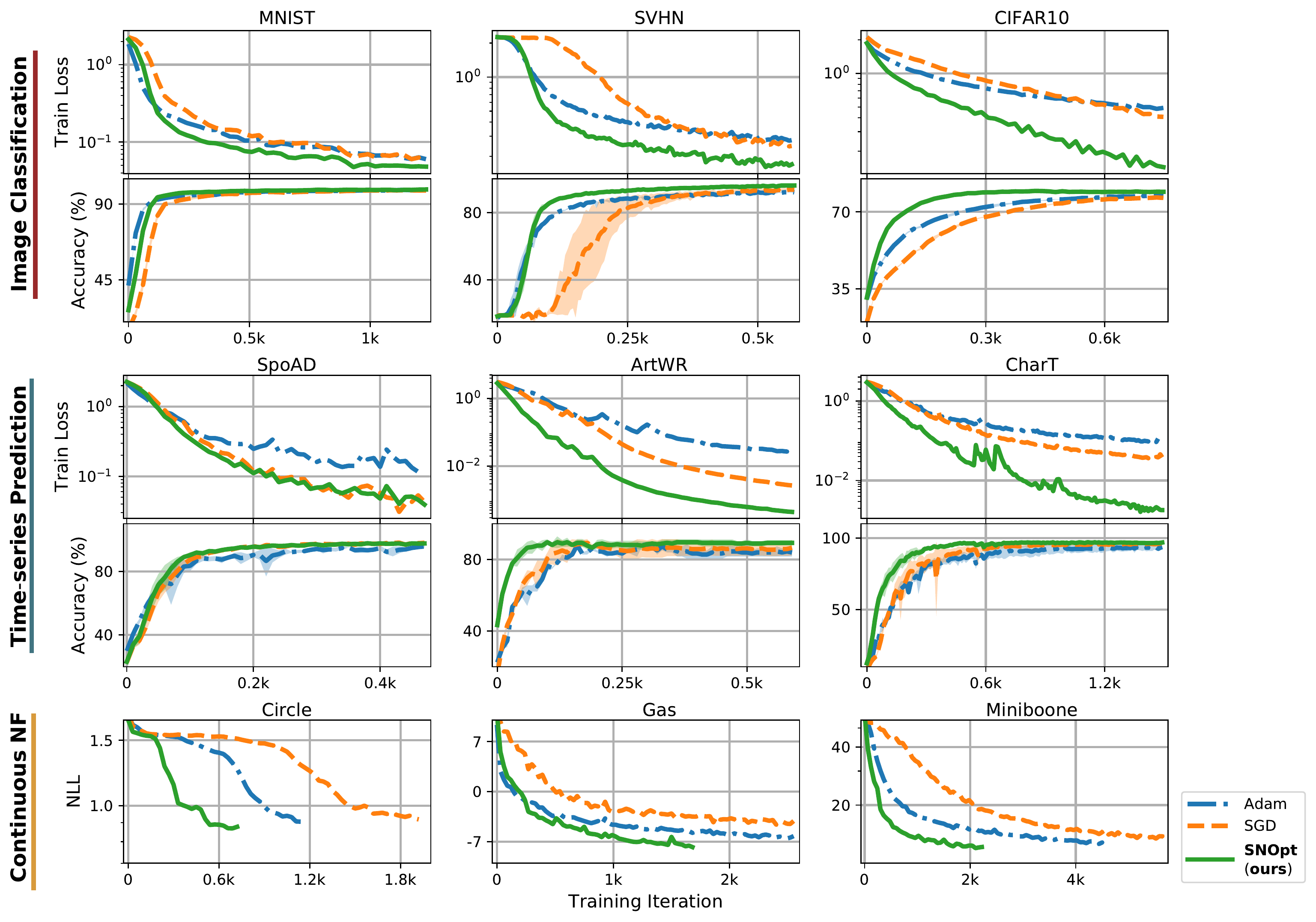}
        \vskip -0.05in
        \caption{
        Optimization performance measured by \emph{iteration updates}
        across 9 datasets, including image (1$^\text{st}$-2$^\text{nd}$ rows) and time-series (3$^\text{rd}$-4$^\text{th}$ rows) classification, and continuous NF (5$^\text{th}$ row).
        Each curve is averaged over 3 random trials.
        }
        \label{fig:convergence-full2}
        \end{center}
    \end{minipage}
    \vskip -0.2in
\end{figure}

\textbf{Convergence on all datasets.}
Figures~\ref{fig:convergence-full} and \ref{fig:convergence-full2} report the training curves of all datasets measured either by the wall-clock time or training iteration.

\textbf{Comparison with first-order methods that handle numerical errors.}
    Table~\ref{table:mali-accu} and \ref{table:mali-runtime} report the performance difference
    between vanilla first-order methods (\eg Adam, SGD), first-order methods equipped with error-handling modules (specifically MALI \citep{zhuang2021mali}), and our SNOpt.
    While MALI does improve the accuracies of vanilla first-order methods at the cost of extra per-iteration runtime (roughly 3 times longer), our method achieves highest accuracy among all optimization methods and retains a comparable runtime compared to \eg vanilla Adam.

\begin{figure}[H]
  \begin{minipage}{\textwidth}
    \vskip -0.05in
    \centering
    \captionsetup{type=table}
    \caption{Test-time performance (accuracies \%) w.r.t. different optimization methods }
    \vskip -0.03in
    \centering
    \begin{tabular}{lccccc}
        \toprule
        & {Adam}  & {Adam + MALI}
        & {SGD}  & {SGD + MALI}
        & \textbf{SNOpt} \\
        \midrule
        SVHN    & 91.92  &  91.98  &  93.34  &  94.33  &  \textbf{95.77} \\[2pt]
        CIFAR10 & 77.41  &  77.70  &  76.42  &  76.41  &  \textbf{79.11} \\
        \bottomrule
    \end{tabular} \label{table:mali-accu}
  \end{minipage}
  \vskip 0.15in
  \begin{minipage}{\textwidth}
    \centering
    \captionsetup{type=table}
    \caption{Per-iteration runtime (seconds) w.r.t. different optimization methods }
    \vskip -0.03in
    \centering
    \begin{tabular}{lccccc}
        \toprule
        & {Adam}  & {Adam + MALI}
        & {SGD}  & {SGD + MALI}
        & \textbf{SNOpt} \\
        \midrule
        SVHN    & 0.78  &  2.31  &  0.81  &  1.28  &  \textbf{0.68} \\[2pt]
        CIFAR10 & \textbf{0.17}  &  0.55  &  \textbf{0.17}  &  0.23  &  0.20 \\
        \bottomrule
    \end{tabular} \label{table:mali-runtime}
  \end{minipage}
  \vskip -0.1in
\end{figure}

\textbf{Comparison with LBFGS.}
Table~\ref{table:lbfgs} reports various evaluational metrics between LBFGS and our {SNOpt} on training MNIST. First, notice that our method achieves superior final accuracy compared to LBFGS. Secondly, while both methods are able to converge to a reasonable accuracy (90\%) within similar iterations, our method runs 5 times faster than LBFGS per iteration; hence converges much faster in wall-clock time.
In practice, we observe that LBFGS can exhibit unstable training without careful tuning on the hyper-parameter of Neural ODEs, \eg the type of ODE solver and tolerance.

\begin{figure}[H]
  \begin{minipage}{\textwidth}
    \centering
    \captionsetup{type=table}
    \caption{Comparison between LBFGS and our SNOpt on training MNIST}
    \vskip -0.03in
    \centering
    \begin{tabular}{lcccc}
        \toprule
        & {Accuracy (\%)}  & {Runtime (sec/itr)}
        & {Iterations to Accu. 90\%}  & {Time to Accu. 90\%} \\
        \midrule
        LBFGS          & 92.76  &  0.75  &  111 steps  &  2 min 57 s \\[2pt]
        \textbf{SNOpt} & 98.99  &  0.15  &  105 steps  &  18 s \\
        \bottomrule
    \end{tabular} \label{table:lbfgs}
  \end{minipage}
\end{figure}

\textbf{Results with different ODE solver (\texttt{implicit adams}).}
Table~\ref{table:iadams} reports the test-time performance when we switch the ODE solver from
\texttt{dopri5} to \texttt{implicit adams}.
The result shows that our method retains the same leading position as appeared in Table~\ref{table:accu}, and the relative performance between optimizers also remains unchanged.

\begin{figure}[H]
  \vskip -0.05in
  \centering
  \setlength\tabcolsep{4.5pt}
  \captionsetup{type=table}
  \caption{Test-time performance using ``\texttt{implicit adams}'' ODE solver: accuracies for {\color{label1}image} and {\color{label3}time-series} datasets; NLL for {\color{label2}CNF} datasets}
  \vskip -0.03in
  \centering
  \begin{tabular}{lccccccccc}
    \toprule
    & {MNIST}  & {SVHN} & {CIFAR10}
    & {SpoAD}  & {ArtWR} & {CharT}
    & Circle & {Gas}  & {Miniboone} \\
    \addlinespace[-0.25em]\arrayrulecolor{label1}
    \cmidrule[1pt](lr){2-4}\corcmidrule\arrayrulecolor{label3}%
    \cmidrule[1pt](lr){5-7}\corcmidrule\arrayrulecolor{label2}%
    \cmidrule[1pt](lr){8-10}\corcmidrule
    \addlinespace[0.4em]\arrayrulecolor{black}
    \midrule
    Adam & 98.86  &  91.76  &  77.22  &  95.33  &  86.28  &  88.83  &  0.90  &  -6.51  &  13.29 \\[2pt]
    SGD  & 98.71  &  94.19  &  76.48  &  \textbf{97.80}  &  87.05  &  95.38  &  0.93  &  -4.69  &  13.77 \\[1pt]
    \midrule
    \textbf{SNOpt}
                  & \textbf{98.95} &  \textbf{95.76} &  \textbf{79.00}
                  & 97.45          &  \textbf{89.50} &  \textbf{97.17}
                  & \textbf{0.86}  &  \textbf{-7.41} &  \textbf{12.37} \\
    \bottomrule
  \end{tabular} \label{table:iadams}
\end{figure}

\textbf{Comparison with discrete-time residual networks.}
Table~\ref{table:resnet}
reports the training results where we replace the Neural ODEs with
standard (\ie discrete-time) residual layers,
$\rvx_{k+1} =  \rvx_k + F(\rvx_k, \theta)$.
Since ODE systems can be made invariant w.r.t. time rescaling
(\eg consider
$\fracdiff{x}{t} = F(t, x, \theta)$ and $\tau = c t $,
then $\fracdiff{x}{\tau} = \frac{1}{c} F(\frac{\tau}{c}, x, \theta)$
will give the same trajectory $x(t) = x(\frac{\tau}{c})$),
the results of these residual networks provide
a performance validation for our joint optimization of $t_1$ and $\theta$.
Comparing
Table~\ref{table:resnet} and \ref{table:adap-t1} on training CIFAR10,
we indeed find that
SNOpt is able to reach the similar performance (77.82\% \textit{vs.} 77.87\%) of the residual network, whereas the ASM baseline gives only 76.61\%, which is 1\% lower.

\begin{figure}[H]
  \begin{minipage}{\textwidth}
    \centering
    \captionsetup{type=table}
    \caption{Accuracies (\%) of residual networks trained with Adam or SGD }
    \vskip -0.03in
    \centering
    \begin{tabular}{lccc}
        \toprule
        & MNIST & SVHN & CIFAR10 \\
        \midrule
        resnet + Adam  & 98.75 $\pm$ 0.21  &  97.28 $\pm$ 0.37  &   77.87 $\pm$ 0.44 \\[2pt]
        \bottomrule
    \end{tabular} \label{table:resnet}
  \end{minipage}
\end{figure}

\textbf{Batch size analysis.}
Table~\ref{table:batch} provides results on image classification when we enlarge the batch size by the factor of 4 (\ie 128 $\rightarrow$ 512). It is clear that our method retains the same leading position with a comparatively smaller variance. We also note that while enlarging batch size increases the memory for all methods, the ratio between our method and first-order baselines does not scale w.r.t. this hyper-parameter. Hence, just as enlarging batch size may accelerate first-order training, it can equally improve our second-order training. In fact, a (reasonably) larger batch size has a side benefit for second-order methods as it helps stabilize the preconditioned matrices, \ie $\bar{\mA}_n$ and $\bar{\mB}_n$ in (\ref{eq:layer-kfac}), throughout the stochastic training (note that too large batch size can still hinder training \citep{keskar2016large}).

\begin{figure}[H]
  \begin{minipage}{\textwidth}
    \centering
    \captionsetup{type=table}
    \caption{Accuracies (\%) when using larger (128 $\rightarrow$ 512) batch sizes}
    \vskip -0.03in
    \centering
    \begin{tabular}{lccc}
        \toprule
        & MNIST & SVHN & CIFAR10 \\
        \midrule
        Adam  & 99.14 $\pm$ 0.12  &  94.19 $\pm$ 0.18  & 77.57 $\pm$ 0.30 \\[2pt]
        SGD   & 98.92 $\pm$ 0.08  &  95.67 $\pm$ 0.48  & 76.66 $\pm$ 0.29 \\[1pt]
        \midrule
        \textbf{SNOpt}
              & \textbf{99.18} $\pm$ \textbf{0.07}  &  \textbf{98.00} $\pm$ \textbf{0.12}  & \textbf{80.03} $\pm$ \textbf{0.10} \\
        \bottomrule
    \end{tabular} \label{table:batch}
  \end{minipage}
\end{figure}